\pdfoutput=1

\documentclass[11pt]{article}
\usepackage[pagebackref]{hyperref}

\usepackage[]{acl2021}

\usepackage{times}
\usepackage{latexsym}

\usepackage[T1]{fontenc}

\usepackage[utf8]{inputenc}

\usepackage{microtype}
\usepackage[LGR,T1]{fontenc}
\usepackage[utf8]{inputenc}   %
\newcommand{\textgreek}[1]{\begingroup\fontencoding{LGR}\selectfont#1\endgroup}

\usepackage{amsmath,bm}
\usepackage{booktabs}
\usepackage{multirow}
\usepackage{graphicx}
\graphicspath{ {images/} }
\usepackage{placeins}
\usepackage{enumitem}

\usepackage{subcaption}
\usepackage{stfloats}

\usepackage[
    textsize=tiny,
    textcolor=black,
    linecolor=black,
    bordercolor=black,
    backgroundcolor=pink,
]{todonotes}

\title{Automatic Evaluation and Analysis of Idioms \\in Neural Machine Translation}

\author{Christos Baziotis\thanks{~~This work was done during an internship at Amazon.} \\
  University of Edinburgh \\
  \texttt{c.baziotis@ed.ac.uk} \\
  \And
  Prashant Mathur \\
  Amazon AI \\
  \texttt{pramathu@amazon.com} \\
  \And 
  Eva Hasler \\
  Amazon AI \\
  \texttt{ehasler@amazon.com}}

\date{}

\begin{document}
\maketitle
\begin{abstract}

A major open problem in neural machine translation (NMT) 
is the translation of idiomatic expressions, 
such as ``\textit{under the weather}''.
The meaning of these expressions is not composed by the meaning of their constituent words,
and NMT models tend to translate them literally (i.e., word-by-word),
which leads to confusing and nonsensical translations.
Research on idioms in NMT is limited 
and obstructed by the absence of automatic methods 
for quantifying these errors.
In this work, first, we propose a novel metric for \textit{automatically} 
measuring the frequency of literal translation errors without human involvement.
Equipped with this metric, 
we present \textit{controlled} translation experiments with models trained 
in different conditions (with/without the test-set idioms)
and across a wide range of (global and targeted) metrics and test sets.
We explore the role of monolingual pretraining
and find that it yields substantial \textit{targeted} improvements,
even without observing any translation examples of the test-set idioms. 
In our analysis, we probe the role of idiom context.
We find that the randomly initialized models are more local or ``myopic'' 
as they are relatively unaffected by variations of the idiom context,
unlike the pretrained ones.

\end{abstract}

\section{Introduction}
\label{sec:intro}

Neural machine translation (NMT;~\citealt{sutskever2014sequence, Bahdanau2014,vaswani2017attention}) 
struggles with the translation of rare multi-word expressions (MWE)~\cite{koehn-knowles-2017-six}.
Non-compositional phrases, such as idioms (e.g., ``piece of cake''), are one of the most challenging types of MWEs,
because their meaning is figurative and cannot be derived from the meaning of their constituents~\cite{Nunberg1994-NUNI,liu2017idioms}.
NMT models tend to translate these expressions
literally (i.e., word-by-word), which leads to erroneous translations.
In this paper, our focus is on the translation of idiomatic expressions,
in contrast to most prior work, 
which is subsumed under MWEs in general~\cite{constant-etal-2017-survey, mwe-2021-multiword}.

The absence of \textit{targeted} and \textit{automatic} evaluation is a major obstacle to advances in idiom translation.
Global metrics, such as BLEU~\cite{papineni-etal-2002-bleu}
consider the full translation, 
and thus, the effects of idiom translation are overshadowed.
Previous efforts on targeted evaluation
isolate the idiom translation using word alignments~\cite{fadaee-etal-2018-examining} or word edit distance~\cite{zaninello-birch-2020-multiword}.
These approaches measure the accuracy of idiom translation but do not account for literal translation errors.
\citet{shao-etal-2018-evaluating} proposed a method 
for estimating the frequency of such errors, 
but it requires the creation of language-specific hand-crafted lists  (i.e., blocklists) with words that correspond to literal translation errors.

In this work\footnotemark, we present a study of idioms in NMT, 
with the goal of facilitating future research in this direction. 
First, we propose a novel metric for the \textit{automatic} evaluation of literal translation errors (LitTER),
that does not require any hand-crafted blocklists.
We incorporate LitTER, which complements alignment-based metrics~\cite{fadaee-etal-2018-examining}
into a unified targeted evaluation framework.

Next, we present translation experiments in a controlled setting,
by using different training splits to test models under different conditions (e.g., zero-shot).
To improve idiom translation we leverage monolingual data, 
which are more abundant than parallel
and contain idioms in higher frequencies and more diverse contexts.
We exploit monolingual data via pretraining (mBART;~\citealt{liu-etal-2020-multilingual-denoising}),
which is a generic and task-agnostic approach, 
unlike prior work that considers ad-hoc solutions~\cite{fadaee-etal-2018-examining, zaninello-birch-2020-multiword}.
We find that monolingual pretraining yields strong targeted gains, 
even when models have not seen any translation examples of the test idioms.

\footnotetext{Code and data in \href{https://github.com/amazon-research/idiom-mt}{github.com/amazon-research/idiom-mt}}

We also present an extensive analysis of how different models translate idioms.
Specifically, we use a series of probing methods
that encode idioms within different contexts~\cite{garcia-etal-2021-probing,yu-ettinger-2020-assessing}, 
and measure how this affects the translation outputs and the decoder distributions. 
We find that the randomly initialized models are more ``myopic'' compared to the pre-trained ones, 
as they are relatively unaffected when we vary the idiom context.
Our contributions are:
\setlist[enumerate]{leftmargin=17pt}
\begin{enumerate}
[topsep=3pt,itemsep=3pt,partopsep=0pt, parsep=0pt]
    \item We propose LitTER (\S\ref{sec:eval-litter}), 
    a novel metric for measuring the frequency of literal translation errors,
    and embed it into a framework (\S\ref{sec:eval}) for \textit{automatic} and \textit{targeted} evaluation of idiom translation, complementing prior work.
    \item We present translation results (\S\ref{sec:results}) in a \textit{controlled} setting 
    and across a wide range of metrics.
    We find that pre-training on monolingual data yields substantial \textit{targeted} improvements.
    \item We present an extensive \textit{analysis} (\S\ref{sec:analysis}) with a series of probes, showing how context affects idiom translation.
    We find that models are more uncertain when translating idioms
    and that pretraining makes models more contextual. %
\end{enumerate}

\section{Automatic Targeted Evaluation}
\label{sec:eval}

\subsection{Literal Translation Error Rate (LitTER)}
\label{sec:eval-litter}

We propose literal translation error rate (LitTER),
a novel metric of the frequency of literal translation errors made by a model.
A literal translation error occurs if any of the words of a span in the source sentence has been \textit{wrongly} translated literally in the target language.
Our metric is inspired by the method of~\citet{shao-etal-2018-evaluating}
which identifies possible literal translation errors, 
by checking if a translation output contains any blocklisted words. 
While this method is effective at capturing these errors, it relies on hand-crafted blocklists.
We overcome this limitation by automatically creating word blocklists for a given expression. 

Our method, is based on two key ideas.
First, we use bilingual word dictionaries\footnotemark, 
which are relatively easy to obtain, 
to translate the words of an annotated source span into the target language, 
and produce blocklists with candidate literal translation errors.
Then, we use the reference translations to filter the blocklists
by removing those words that occur in the reference. 
This avoids triggering the blocklist when the correct translation is literal.

\footnotetext{In this work we use the MUSE~\cite{lample2018word}.
}

\begin{figure}[t]
    \centering
    \includegraphics[width=1\columnwidth]{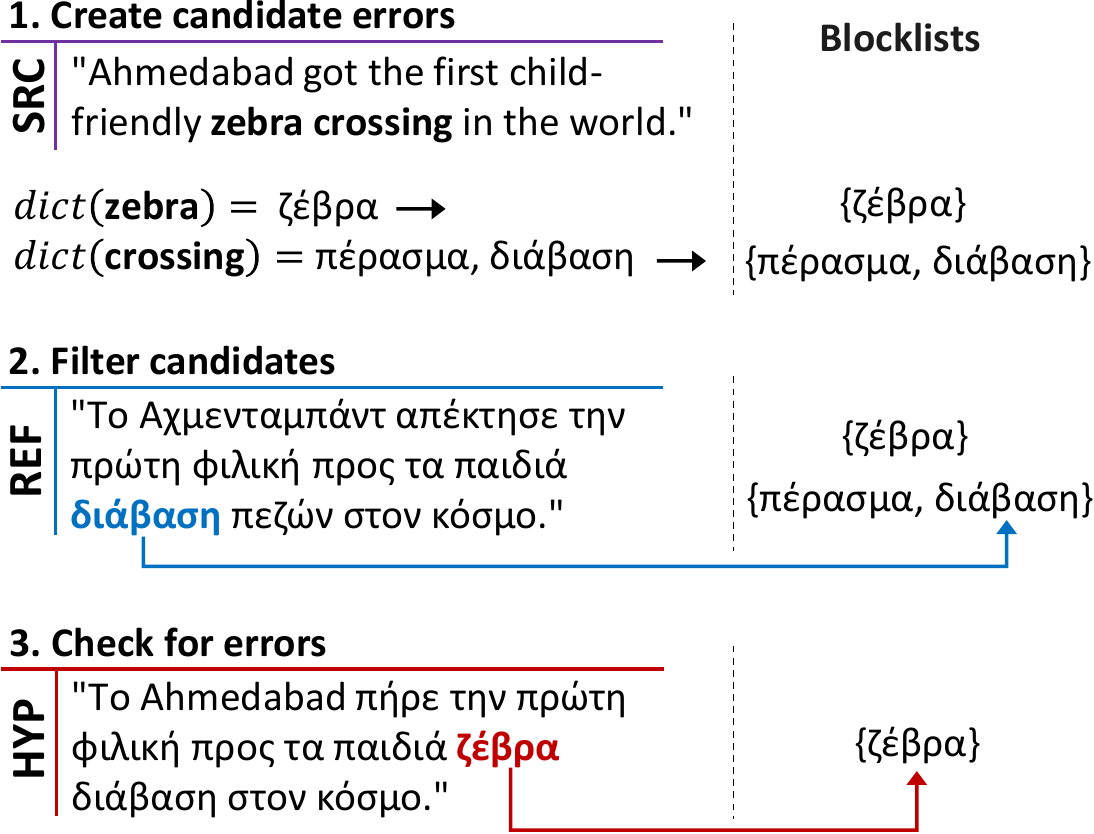}
    \caption{Overview of the algorithm for the Literal Translation Error Rate (LitTER).
    For each sentence, we first produce candidate literal translation errors (blocklist), using all the word translations of the source idiom words. Then, we filter the candidates in the blocklist by looking at the reference. Finally, we check if the hypothesis triggers the remaining words in the blocklist.}
    \label{fig:litter}
\end{figure}

\paragraph{Algorithm}
\setlist[enumerate]{leftmargin=20pt}
\begin{enumerate}
[topsep=5pt,itemsep=5pt,partopsep=0pt, parsep=0pt]
    \item Select from the source text
    the list of words $\bm{s}= \langle s_1, s_2, ..., s_N \rangle$ that belong to the annotated expression (i.e., idiom).
    \item For each word $s_i$, 
    obtain all its word translation(s) in the target language
    using a bilingual word dictionary 
    and add them to a blocklist $b_i = \langle t_1, t_2, \ldots, t_M \rangle$, creating a candidate \textit{list} of blocklists
    $\bm{B_\bm{s}}= \langle b_1, b_2, ..., b_N \rangle$.\footnote{In practice, $t_1, t_2, \ldots, t_M$ in a blocklist are synonyms of each other as they are translations of the same source word.}
    \item For each word in the reference (R), 
    search if it occurs in any of the blocklists $b_i$.
    If so, remove the corresponding blocklist $b_i$ 
    from $\bm{B_\bm{s}}$ to avoid false positives.
    For example in Figure~\ref{fig:litter},
    where words \textgreek{διάβαση} and \textgreek{πέρασμα} are synonyms, if we remove only \textgreek{διάβαση} but leave \textgreek{πέρασμα} as a blocklisted word and a model generates it in its translation,
    this will \textit{wrongly} trigger a literal translation error.
    \item Check if the hypothesis contains any blocklisted words. 
    If it does, then we mark this hypothesis as having a literal translation error.
\end{enumerate}

\noindent The final score is the percentage of translations that trigger the blocklist. 
As LitTER requires source-side annotations, 
we collect test data with idioms on the source side and 
annotate the spans where they occur (\S\ref{sec:experiments-data}). 
Appendix~\ref{sec:app-litter} shows examples of LitTER evaluating real sentences in our data.

\subsection{Alignment-based Evaluation}
\label{sec:eval-apt}

To measure idiom translation accuracy,
we use Alignment-based Phrase Translation Evaluation (APT-Eval),
by extending \citet{fadaee-etal-2018-examining} with subword-level metrics.
APT-Eval uses word alignments to find the words in the hypothesis and reference sentences, respectively, 
that align with the annotated idiom source span, 
and then compares the retrieved matches to each other. 
We consider two evaluation metrics.
First, we use \textit{unigram precision}, 
that measures the ratio of words in the reference spans that occur in the hypothesis spans, 
as in~\citet{fadaee-etal-2018-examining}.
We also use ChrF~\cite{popovic-2015-chrf}, that measures character n-gram overlap.

\paragraph{LitTER vs. APT-Eval}
While APT-Eval is a targeted evaluation metric, it only measures translation accuracy.
This means that given an inaccurate translation, it is impossible to measure whether it has a literal translation error.
LitTER, however, quantifies this particular issue that affects NMT.

\subsection{Handling Idiom Frequency Imbalances}
\label{sec:eval-macro}

Different idioms have significantly different frequencies (Appendix~\ref{sec:app-idiom-stats}).
However, prior work has overlooked this fact~\cite{zaninello-birch-2020-multiword, fadaee-etal-2018-examining, shao-etal-2018-evaluating, riktersbojar2017}.
Thus, over-represented idioms can skew the reported results 
and favour models that have overfitted on them.
To address this, we report all of our targeted evaluation results (i.e., LitTER, APT-Eval) by macro-averaging over idioms:
\begin{align}
    \bm{E}(\theta) = \frac{1}{|\bm{L}|}\sum_{j=1}^{|\bm{L}|} 
    \frac{1}{|\bm{L_j}|} \sum_{i=1}^{|\bm{P}|}\bm{M}(\theta(s_{i}), t_{i})
\end{align}
\noindent where $\bm{L}$ denotes the set of distinct idioms in a test set and $\bm{P} = \{\langle s_i, t_i \rangle | L_j \in \langle s_i, t_i\rangle\}$ denotes the set of sentence pairs containing the idiom $L_j$. 
The model is denoted by $\theta$ and the translation of $x$ by $\theta(x)$. 
We first compute the average score for the test pairs of each idiom with a given metric $\bm{M}$, and then average these values to produce $\bm{E}$.

\section{Experiments}
\label{sec:experiments}

\subsection{Data and Training Splits} 
\label{sec:experiments-data}

We present experiments on en$\rightarrow$fr and en$\rightarrow$es data.
For each language pair, 
we concatenate the data from Europarl~v7\footnotemark~\cite{koehn-2005-europarl}, 
\footnotetext{\href{https://www.statmt.org/europarl/}{www.statmt.org/europarl/}}
part of the WMT news translation task~\cite{bojar-etal-2014-findings}, 
and from TED talk transcripts released as part of IWSLT 2017 shared task\footnotemark~\cite{cettolo2017overview}.

\footnotetext{\href{https://sites.google.com/site/iwsltevaluation2017/TED-tasks}{sites.google.com/site/iwsltevaluation2017/TED-tasks}}

\paragraph{Idiom Data}
We split the parallel data into regular and idiom data using a pattern-matching tool 
that we developed.
Our tool takes as input a list of idioms and extracts sentences from a corpus containing these idioms.
We also annotate the span in which each idiom occurs within a sentence, to enable the targeted evaluation metrics. 
This approach is similar to \citet{fadaee-etal-2018-examining}, 
but we build our tool on top of Spacy's~\cite{spacy2} rule-based matching engine. 
For each phrase in the input list, 
we automatically create pattern-matching rules that capture complex variations of a given phrase.
See Appendix~\ref{sec:app-idiom-data} for details.

In this work, we use a list of 225 English idioms, 
that we manually collected and plan to make it publicly available. 
We feed this list into our pattern-matching tool, 
and extract (and annotate) translation pairs that contain an idiom on the source side.
The regular data are used \textit{only} for training. 
The idiom data are further divided into the \textit{idiom-train} and \textit{idiom-test} sets. 
For each idiom (e.g., ``under the weather'') in our original idiom data, we put half of its sentence pairs to the idiom-train and the other half to the idiom-test sets,
to obtain a balanced distribution. 
We discard sentences with idioms that occur only once.
We conduct controlled experiments, in the following testing conditions:
\setlist[itemize]{leftmargin=20pt}
\begin{itemize}
[topsep=3pt,itemsep=4pt,partopsep=0pt, parsep=0pt]

\item{\textbf{Zero}}: training data includes only regular parallel data,
and we measure how models perform on unseen idioms at test time.

\item{\textbf{Joint}}: training data includes the regular and idiom-train data,
and we measure how models perform on idioms observed (in a different context) in training data.

\item{\textbf{Upsampling}}: same as the joint split, but we up-sample the idiom-train data $N$ times. This setting measures whether it is necessary to up-sample the targeted training data (idiom-train) to achieve better translation quality of idioms.

\end{itemize}

\begingroup
\setlength{\tabcolsep}{4.5pt} %
\renewcommand{\arraystretch}{0.9} %
\begin{table}[t]
  \small
  \centering
  \begin{tabular}{lrr}
    \toprule[1.5pt]
    \textbf{Data}                & \textbf{en$\rightarrow$fr} & \textbf{en$\rightarrow$es} \\
    \midrule
    Europarl            & 2,007,723         & 1,965,734         \\
    IWSLT               & 275,085           & 265,625           \\
    Combined (after preprocessing) & 2,155,543         & 2,119,686         \\
    \midrule
    Regular             & 2,152,716         & 2,116,889         \\
    Idiom-train         & 1,327             & 1,312             \\
    \midrule
    Idiom-test          & 1,383             & 1,373             \\
    WMT-test            & 3,003             & 3,000             \\
    IWSLT-test          & 2,632             & 2,502             \\
    \bottomrule[1.5pt]
  \end{tabular}
  \caption{Dataset statistics}
  \label{table:data-statistics}
\end{table}

\endgroup

\paragraph{Evaluation} For development, we use the IWSLT dev-set for each language pair. 
For general purpose translation evaluation, we report results in the WMT newstest14 and IWSLT'17 test sets for en$\rightarrow$fr, and in the WMT newstest13 in particular and IWSLT'17 test sets for en$\rightarrow$es. For the targeted idiom evaluation (i.e., LitTER and APT-Eval) we use the extracted idiom-test data per language pair.
To generate the word alignments for APT-Eval, 
we trained a fast-align~\cite{dyer-etal-2013-simple} model %
on each language-pair's training data.
For decoding, we use beam search with beams of size 5,
and evaluate all models using BLEU~\cite{papineni-etal-2002-bleu} 
computed with Sacre\textsc{bleu}~\cite{post-2018-call}.

\paragraph{Preprocessing} We first filter out sentence pairs with more than 80 words or with length ratio over 1.5. Then, we tokenize the remaining sentences
using sentencepiece\footnotemark (SPM;~\citealt{kudo-richardson-2018-sentencepiece}). 
For the randomly initialized models, we train SPM models with a joint  vocabulary of 60K symbols on the concatenation of the source- and target-side of the regular training data. 
For the mBART fine-tuning experiments, we use the SPM model of mBART (250K symbols).
\footnotetext{We use the \texttt{unigram} model with \texttt{coverage=0.9999}}

\subsection{Models} 
Besides training models from scratch,
we also investigate how pretraining on monolingual data affects idiom translation,
which yields substantial improvements on generic translation quality~\cite{lample2019cross,pmlr-v97-song19d, liu-etal-2020-multilingual-denoising}.
However, it is not obvious if monolingual data can help idiom translation,
as they do not contain any examples with how to translate an idiom from one language into another.

We use mBART~\cite{liu-etal-2020-multilingual-denoising} via finetuning, 
which is pretrained on monolingual data from many languages.
We hypothesize that one way multilingual pre-training can help is by bootstrapping over the source and target language contexts in which idioms occur.
We also consider injecting different types of noise during fine-tuning, 
to corrupt the (encoder or decoder) input context and 
measure the effects on the targeted evaluation metrics.
Specifically, we use source-side word masking and replacement~\cite{baziotis-etal-2021-exploring}, and target-side word-replacement noise~\cite{voita-etal-2021-analyzing}.
In our experiments, ``random'' denotes a randomly initialized model, 
while ``mBART'' stands for using mBART as initialization. 
For noisy finetuning we train the following variants:  
``mBART+mask'' where we mask 10\% of the source tokens,
``mBART+replace (enc)'' where we replace 10\% of the source tokens with random ones, and
``mBART+replace (dec)'' where we replace 10\% of the target tokens with random ones.

\paragraph{Model Configuration} For fair comparison, the randomly initialized models use the same architecture as mBART. Specifically, the models are based on the Transformer architecture, with 12 encoder and decoder layers, 1024 embedding size and 16 self-attention heads. 
Our code is based on the official mBART implementation in Fairseq.

\paragraph{Optimization} We optimized our models using Adam~\cite{kingma2014Adam} with 
$\beta_{1}=0.9, \beta_{2}=0.999$, 
and $\epsilon$=1e-6.
For the random initialization experiments, the models were trained for 140K updates with batches of 24K tokens, using a learning rate of 1e-4 with a linear warm-up of 4K steps, followed by inverted squared decay. 
For the mBART initialization experiments, the models were trained for 140K updates with batches of 12K tokens, using a fixed learning rate of 3e-5 with a linear warm-up of 4K steps. 
In all experiments, we applied dropout of 0.3, attention-dropout of 0.1 and label smoothing of 0.1. 
For model selection, we evaluated each model every 5K updates on the dev set, and selected the one with the best BLEU.

\begingroup
\setlength{\tabcolsep}{7.5pt} %
\renewcommand{\arraystretch}{0.8} %
\begin{table*}[t]
\small
\centering
\begin{tabular}{clrrrrrr}
\toprule[1.5pt]
\multirow{2}[2]{*}{\textbf{Split}}
& \multirow{2}[2]{*}{\textbf{Model (en$\rightarrow$fr)}}
& \multirow{2}{*}{\textbf{LitTER$\downarrow$}} 
& \multicolumn{2}{c}{\textbf{APT-Eval}} 
& \multicolumn{3}{c}{\textbf{Global Evaluation (BLEU$\uparrow$)}} \\
\cmidrule(lr){4-5} \cmidrule(lr){6-8}
                                                                          &                             &                          & UniPrec↑     & ChrF↑    & IWSLT17       & WMT14       & Idiom-test       \\
\midrule
\multirow{5}{*}{zero}                                                     & random                      & 0.563                    & 0.268             & 0.298    & 44.1          & 34.8        & 34.4        \\
                                                                          & mBART                       & 0.484                    & 0.291             & 0.322    & 47.0          & 38.6        & 36.5        \\
                                                                          & mBART +mask                 & 0.478                    & 0.298             & 0.323    & 46.3          & 38.2        & 36.0        \\
                                                                          & mBART +replace (dec)        & 0.519                    & 0.295             & 0.319    & 46.9          & 39.0        & 36.0        \\
                                                                          & mBART +replace (enc)        & 0.365                    & 0.260             & 0.284    & 44.1          & 36.2        & 34.5        \\
\midrule
\multirow{5}{*}{joint}                                                    & random                      & 0.448                    & 0.317             & 0.337    & 44.2          & 34.8        & 35.3        \\
                                                                          & mBART                       & 0.408                    & 0.333             & 0.352    & 46.5          & 38.5        & 37.3        \\
                                                                          & mBART +mask                 & 0.443                    & 0.315             & 0.338    & 46.2          & 38.3        & 37.2        \\
                                                                          & mBART +replace (dec)        & 0.447                    & 0.317             & 0.342    & 46.8          & 38.8        & 37.0        \\
                                                                          & mBART +replace (enc)        & 0.364                    & 0.300             & 0.322    & 44.5          & 36.6        & 35.6        \\
\midrule
\multirow{2}{*}{\begin{tabular}[c]{@{}c@{}}upsample \\ 20x\end{tabular}}  & random                      & 0.371                    & 0.323             & 0.353    & 44.4          & 34.7        & 35.3        \\
                                                                          & mBART                       & 0.289                    & 0.329             & 0.346    & 46.6          & 38.7        & 36.0        \\
\bottomrule[1.5pt]
\end{tabular}
\caption{All of our (en$\rightarrow$fr) translation results (single run), including \textit{generic} and \textit{targeted} evaluation.}
\label{table:enfr-main}
\end{table*}
\endgroup

\subsection{Results}
\label{sec:results}
In this section, for brevity, we discuss a subset of our results,
in particular our experiments in en$\rightarrow$fr.
Results for en$\rightarrow$es are consistent with en$\rightarrow$fr and are included in Appendix~\ref{sec:app-results}.
Table~\ref{table:enfr-main} summarizes all of our main results.
Besides global evaluation using BLEU (\S\ref{sec:results-regular-mt}) on diverse test sets, 
we also consider two targeted evaluation methods (\S\ref{sec:results-targeted}) 
that focus on how the idioms are translated using our idiom-test set.
For the upsampling split, we upsample the idiom-train data 20x. 
We also experimented with 100x upsampling, 
but models started to exhibit overfitting effects (see \S\ref{sec:app-results}, \S\ref{sec:app-analysis}).

\begin{figure*}[t]
\centering
\begin{subfigure}{.48\textwidth}
  \centering
  \includegraphics[width=1\linewidth]{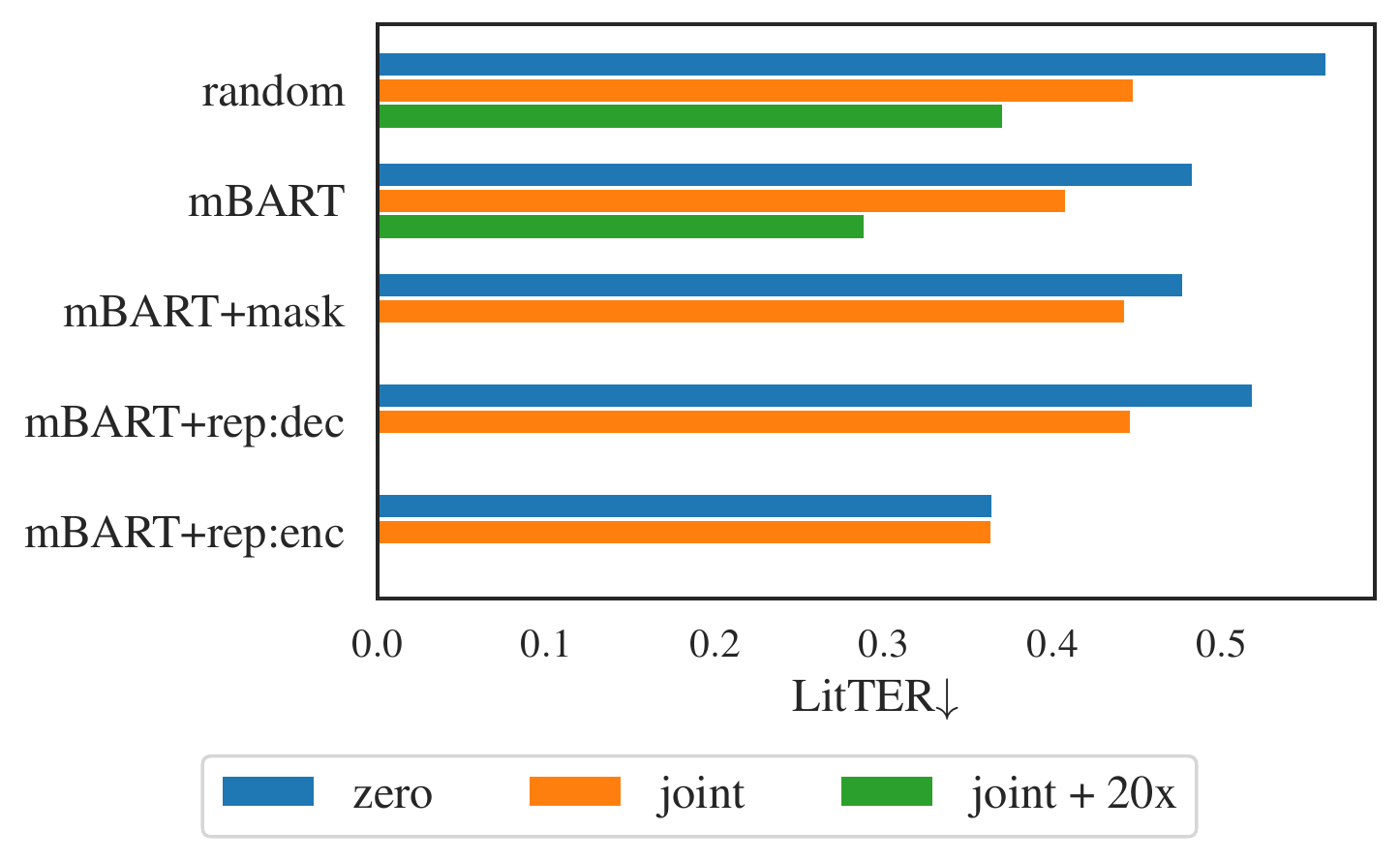}
  \caption{Results on LitTER, which measures how often each model makes literal translation errors.}
  \label{fig:enfr-litter}
\end{subfigure}%
\hfill
\begin{subfigure}{.48\textwidth}
  \centering
  \includegraphics[width=1\linewidth]{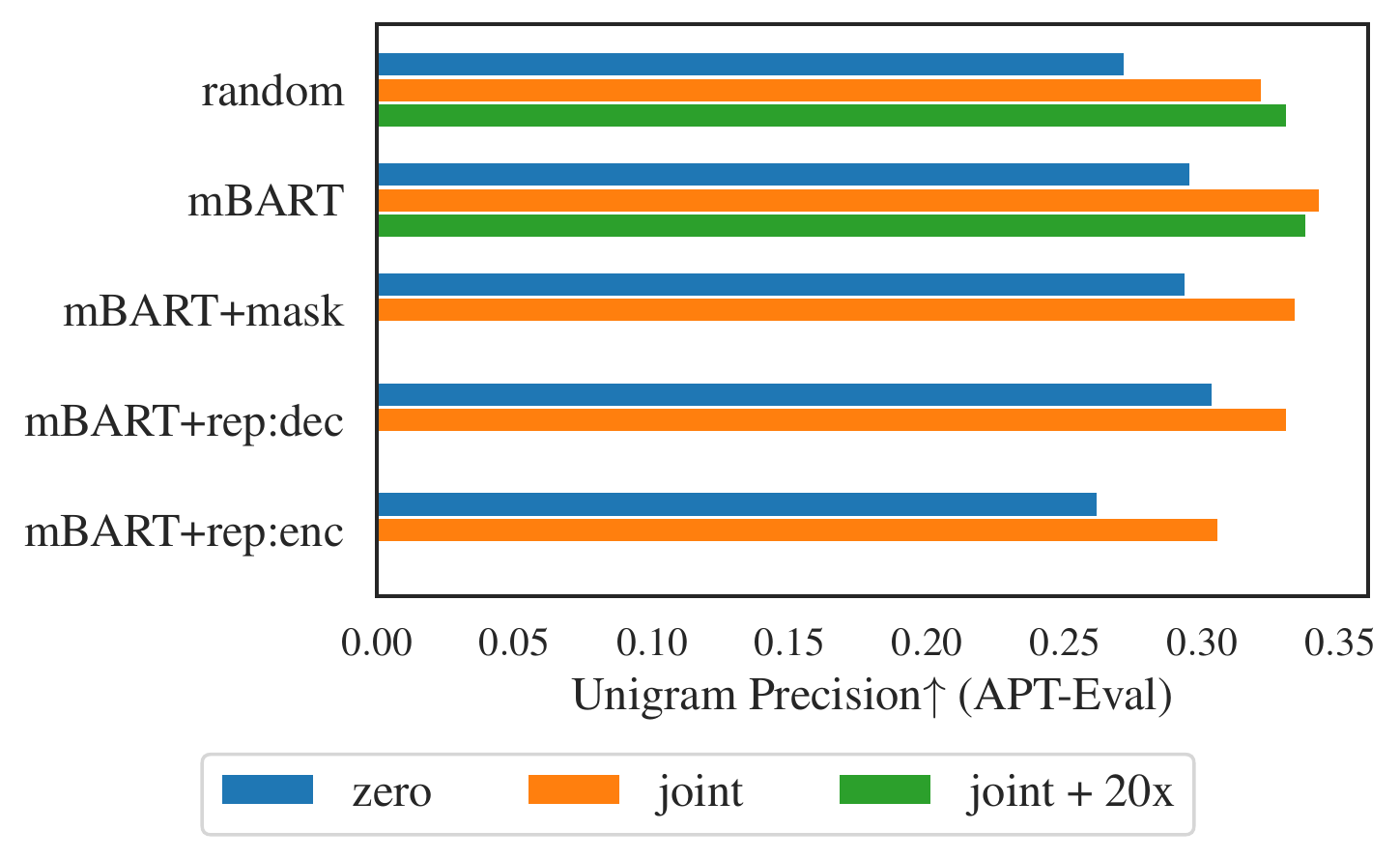}
  \caption{Results on APT-Eval with unigram precision, which compares (the aligned) reference and hypothesis spans.}
  \label{fig:enfr-apt-uprec}
\end{subfigure}
\caption{Results on \textit{targeted} evaluation of idiom translation.
}
\label{fig:en-fr-targeted}
\end{figure*}

\subsubsection{Targeted Evaluation}
\label{sec:results-targeted}

In targeted evaluation, we focus \textit{only} on how models translate the source-side idioms.
We present results on our proposed LitTER metric and on APT-Eval,
which provide different information.
Recall that we macro-average these scores (\S\ref{sec:eval-macro}) to account for imbalances in the idiom frequency.

\paragraph{Literal Translation Errors}
Figure~\ref{fig:enfr-litter}, shows the results on LitTER,
that measures how often models make literal translation errors.
As expected, 
all models produce fewer errors when trained on the joint split compared to the zero split.
Pretraining gives a significant boost,
even on the joint split.
This shows that pretraining helps, 
even though the models have not seen any examples of how to translate the test-set idioms.
Upsampling the idiom-train data helps all models regardless of initialization. 

Each type of noise induces a different behaviour compared to the mBART model. 
Masking yields no effect on the zero split, but increases errors on the joint split. 
\citet{baziotis-etal-2021-exploring}, show that masking promotes copying, which we speculate it could lead to word-by-word translation and increase LitTER.
Decoder-side word replacements yield a similar behaviour in terms of LitTER, 
which we hypothesize could push the decoder to rely more on the encoder, 
therefore encouraging word-by-word translation. 
By contrast, when we add word replacements in the encoder, 
it greatly reduces LitTER in both splits.
This aligns with the findings of \citet{baziotis-etal-2021-exploring}, 
who show that source-side word replacements make the decoder less prone to copying (or “trusting”) the encoder.

\paragraph{Idiom Translation Accuracy}
To estimate how accurately models translate idioms, 
we compare the reference and hypothesis matches that align to the source idiom words. 
Figure~\ref{fig:enfr-apt-uprec}, compares models using unigram precision, 
and the results are consistent across all APT-Eval metrics (Table~\ref{table:enfr-main}).

Similar to the LitTER results, the joint split significantly improves idiom translation accuracy as well.
Again, pretraining outperforms random initialization, even on the joint split.
Upsampling, however, does not yield any consistent improvements.
While source-side word replacements reduce literal translation errors,
they also degrade idiom translation accuracy.
Our hypothesis is that as the decoder becomes less reliant on the encoder,
this induces more hallucinations.

\begin{figure*}[t]
\centering
\begin{subfigure}{.48\textwidth}
  \centering
  \includegraphics[width=1\columnwidth]{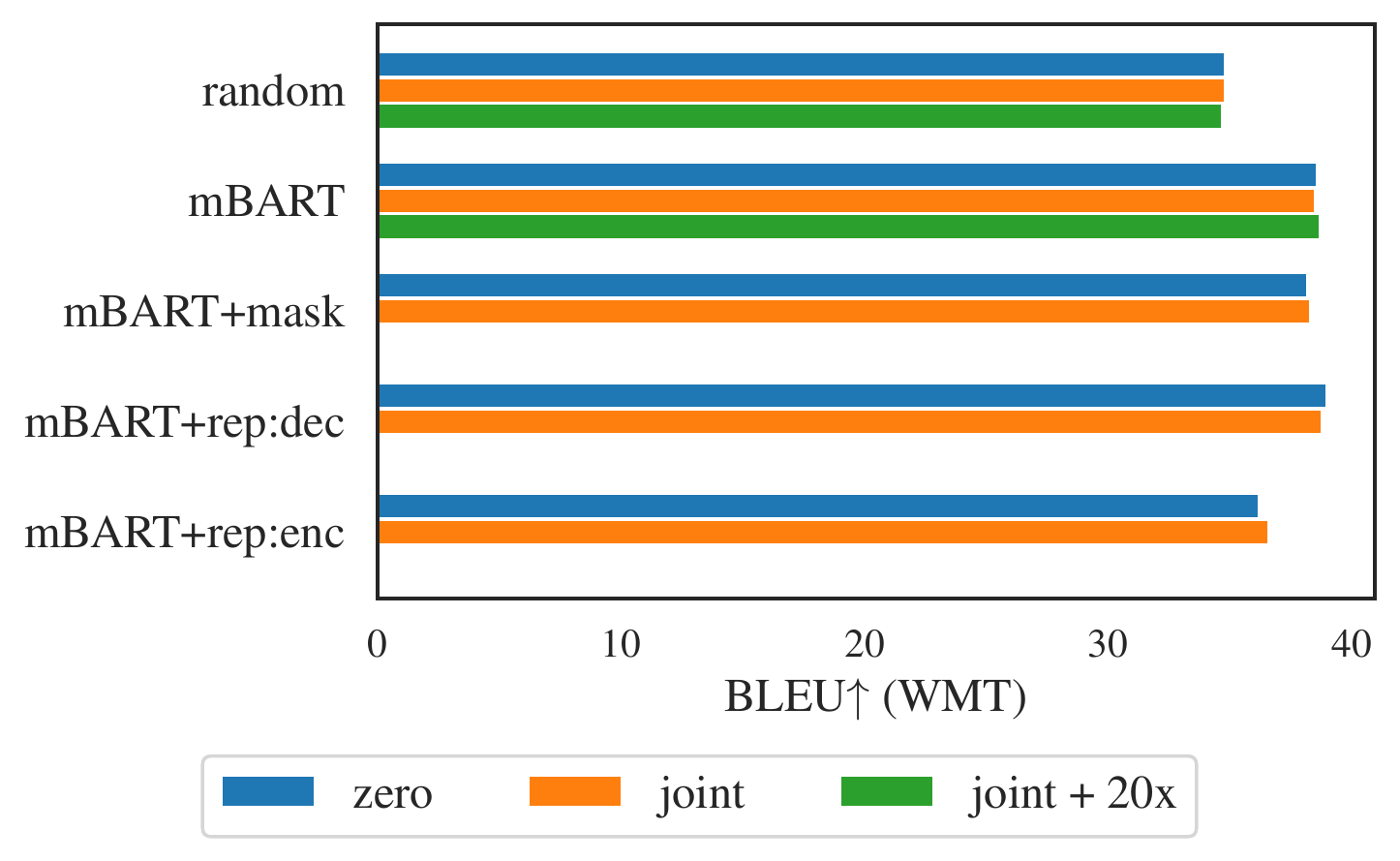}
  \caption{Regular BLEU results on the \textit{generic} WMT14 test set}
  \label{fig:enfr-bleu-wmt}
\end{subfigure}%
\hfill
\begin{subfigure}{.48\textwidth}
  \centering
    \includegraphics[width=1\columnwidth]{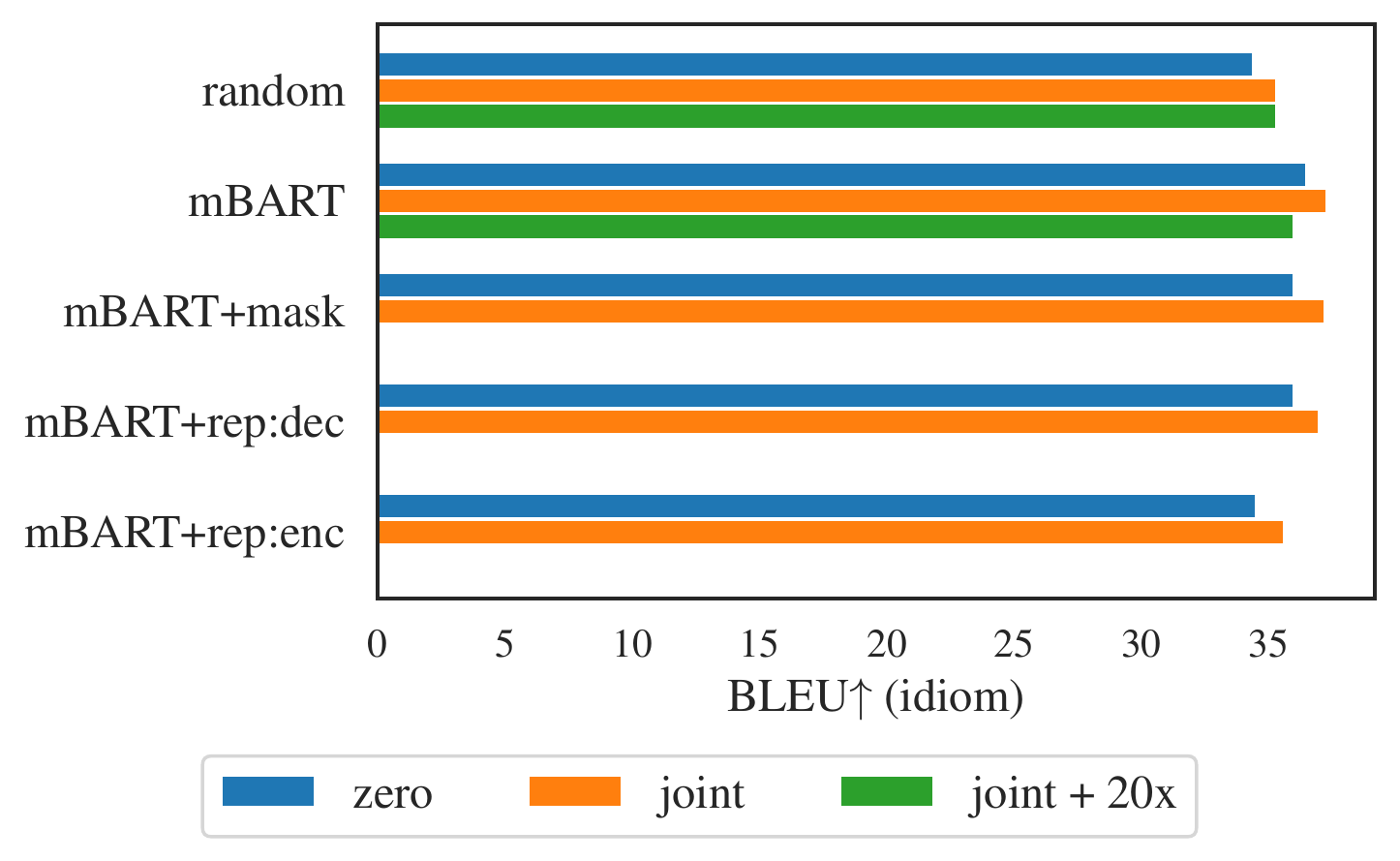}
  \caption{Regular BLEU results on our \textit{idiom-test} set.}
  \label{fig:enfr-bleu-idiom}
\end{subfigure}

\caption{Results on \textit{global} evaluation (BLEU) on different test sets.
}
\label{fig:enfr-bleu}
\end{figure*}

\subsubsection{Global Evaluation}
\label{sec:results-regular-mt}

Here, we discuss how models perform based on global translation evaluation using
BLEU.

\paragraph{General Purpose}

Figure~\ref{fig:enfr-bleu-wmt}, shows the results on the WMT14 test set, but we note that the results are generally consistent with the IWSLT17 test set (Table~\ref{table:enfr-main}).
The mBART intialized models, unsuprisingly, yield significantly better results than random initialization.
As expected, 
there is no measurable difference between splits, not even when upsampling idioms, 
as both IWSLT17 and WMT14 are generic test sets.
Noisy finetuning methods fail to improve results.
Encoder-side word replacements even degrade overall performance, 
which aligns with the hypothesis that they induce hallucinations.

\paragraph{Idiom}
The results on our idiom-test set (Figure~\ref{fig:enfr-bleu-idiom}) show that models perform consistently better when trained on the joint split.
Global evaluation, however, 
considers the full sentences and the impact of idiom translation is overshadowed~\cite{riktersbojar2017},
which can be seen by the very small differences between splits.
This prevents fine-grained comparisons between models
and highlights the need for targeted evaluation.

\section{Analysis}
\label{sec:analysis}

\begin{figure}[t]
    \centering
    \includegraphics[width=1\columnwidth]{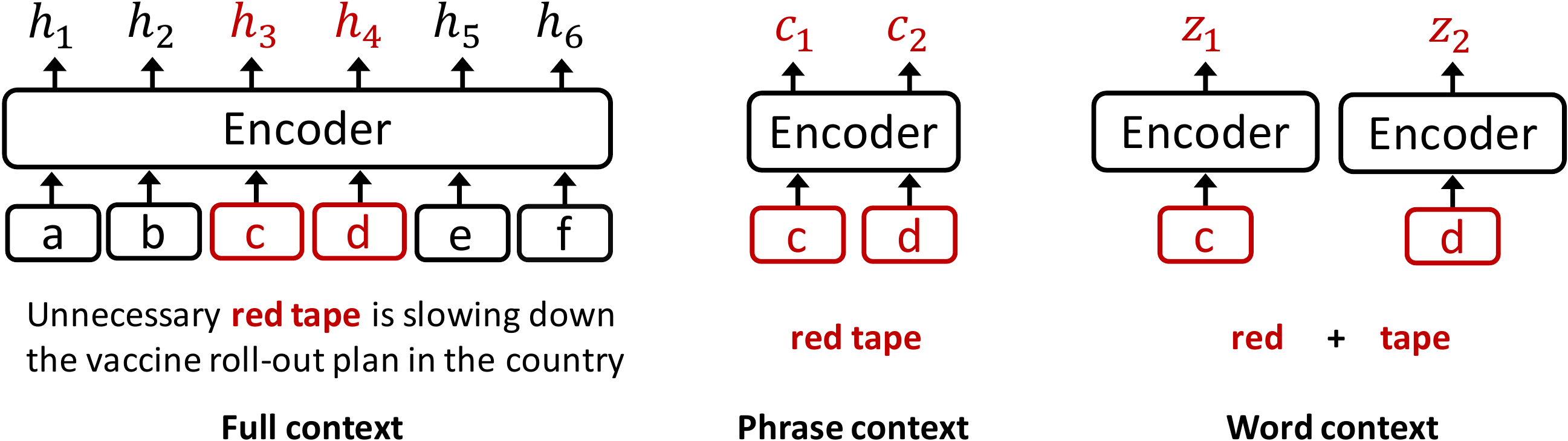}
    \caption{Illustration of how we obtain idiom representations by varying the available context. 
    }
    \label{fig:analysis-overview}
\end{figure}

To further understand how models translate idioms, 
we present an extensive analysis with a series of probes,
focused on the role of idiom context.
Specifically, using the annotations in our idiom-test data, 
we encode the idiom words within different contexts (Figure~\ref{fig:analysis-overview})
and measure how it affects 
the decoder distributions and the translation output. 
We consider (1) \textit{full context}, 
in which we encode the idiom phrase within the whole input sentence,
(2) \textit{phrase-level context}, 
in which we encode the idiom phrase in isolation,
(3) \textit{word-level context}, 
in which we encode each idiom word independently.
Our probes follow a similar approach to prior work that 
evaluate the idiomaticity of (pretrained) Transformer-based models~\cite{garcia-etal-2021-probing,tayyar-madabushi-etal-2021-astitchinlanguagemodels-dataset}
by measuring how idiom representations are affected by their context~\cite{yu-ettinger-2020-assessing},
but we extend these methods to analyze how (pretrained) NMT models translate idioms.
For brevity, we discuss here our most important findings, 
focusing on random-vs-mBART initialization.
However, we include the rest of our results in Appendix~\ref{sec:app-analysis} for completeness.

\begin{figure}[t]
    \centering
    \includegraphics[width=1\columnwidth]{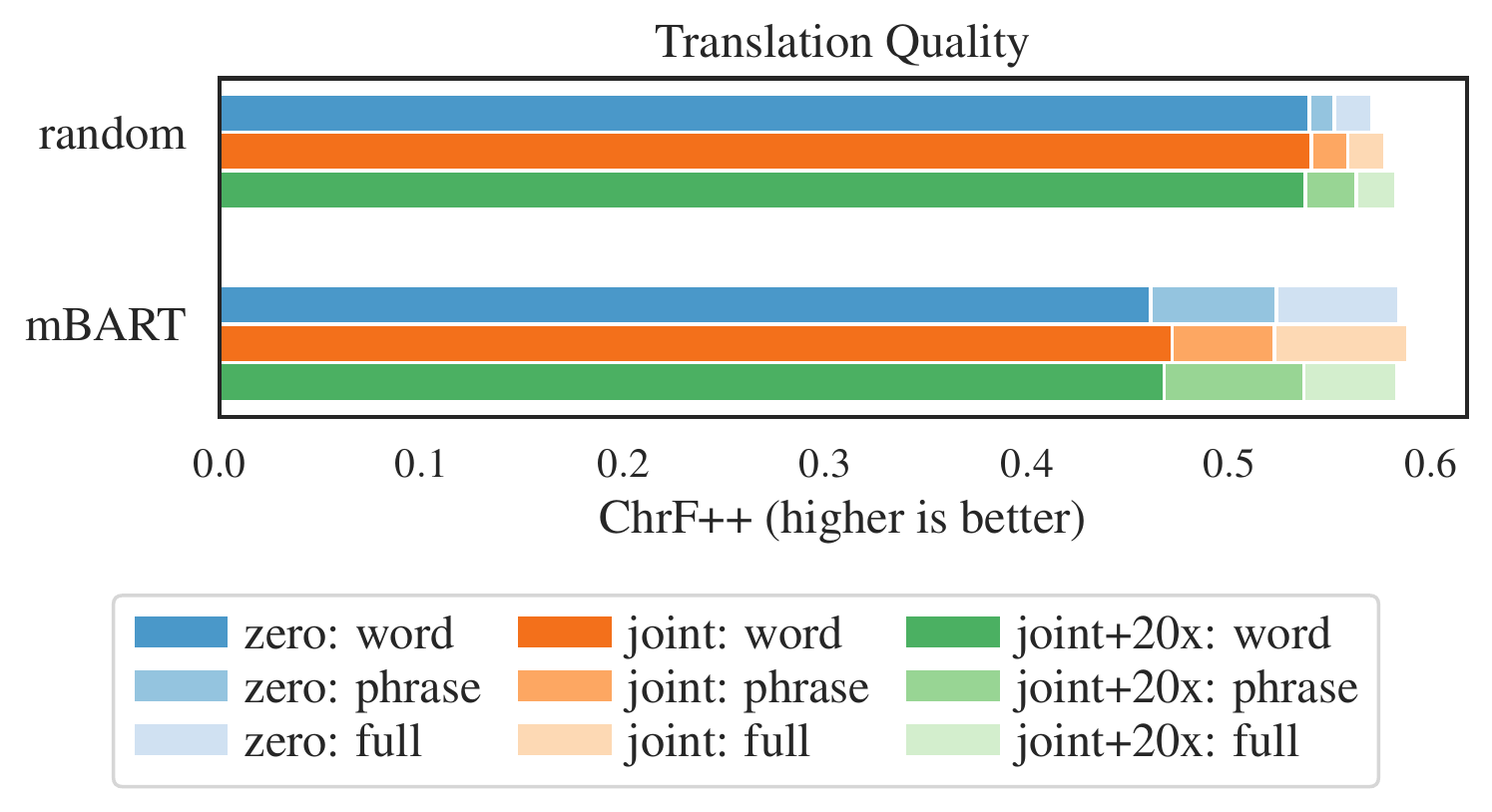}
    \caption{
    Variation in translation quality, measured in ChrF, as we vary the idiom representations.
    The length of each (lighter) bar encodes the difference from its (darker) bar to its left
    (i.e., overlapping bars effect).
    }
    \label{fig:analysis-mt}
\end{figure}

\subsection{Variation in Translation Performance}
\label{sec:analysis-translation}
In this probe, 
we decode (i.e., translate) different encoder representation
and evaluate the samples against the reference (sentence) translations.
Specifically, we first encode each (full) input sentence 
and then replace the encoder representations belonging to idiom words 
with those obtained with different (narrower) contexts.
Figure~\ref{fig:analysis-mt} shows the results using ChrF, rather than BLEU, 
as a metric to capture even small subword-level changes.

Across all models, reducing the context (i.e., darker shades) results in worse translation scores.
This is expected, as by swapping the original (full context) idiom representations with those obtained with word-context, we essentially remove information.
When we use the full-context (i.e, lightest shades) pretraining yields the best results,
but when we reduce the idiom context, 
the pretrained model suffers significantly, unlike the randomly initialized that is barely affected.
This implies that the representations of the randomly initialized model are more local (or ``myopic''), containing information mainly related to the idiom tokens.
By contrast, the representations of the pretrained model are more global, 
and contain information related to the surrounding idiom context~\cite{Brunner2020On}.
Upsampling does not have a strong effect.
\todo[disable]{takeaways: Broader context helps. 
Pretraining yields less myopic models.
Upsampling makes mBART more local?}

\begin{figure}[t]
    \centering
    \includegraphics[width=1\columnwidth]{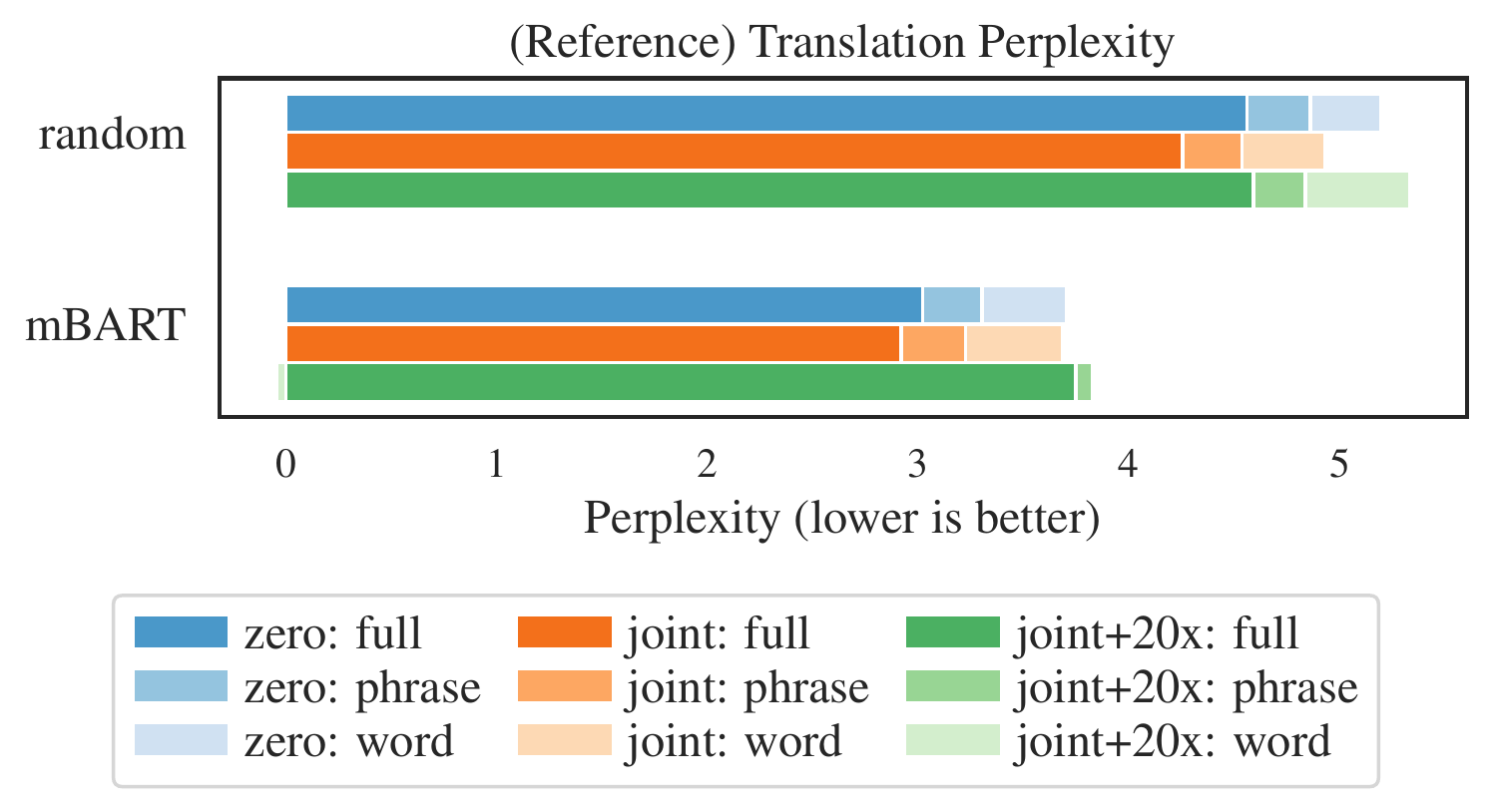}
    \caption{
    Perplexity of the references, as we vary the idiom representations.
    The length of each (lighter) bar encodes the difference from its (darker) bar to its left.
    }
    \label{fig:analysis-perplexity}
\end{figure}

\subsection{Variation in Translation Likelihood}
\label{sec:analysis-likelihood}
In this probe, we vary the encoder idiom representations, as before, 
and measure how this
affects the likelihood of the reference translations.
We score each reference translation by computing its perplexity under the model, 
given each encoder output sequence.
\todo[disable]{takeaways: Broader context helps. 
Upsampling is harmful for general purpose translation.}
Figure~\ref{fig:analysis-perplexity}, shows the results, 
with \textit{lighter} shades corresponding to \textit{narrower} contexts.

Using more context improves~(i.e., lowers) the perplexity of the references across models.
Pretraining endows models with stronger LM capabilities which is probably why it yields generally lower perplexities.
Training on the joint split yields consistent improvements
which are more pronounced in the randomly initialized models.
By contrast, upsampling the idiom-train data yields negative effects, which we attribute to overfitting,
as it makes all other sentences less probable under the model.

\begin{figure}[t]
    \centering
    \includegraphics[width=1\columnwidth]{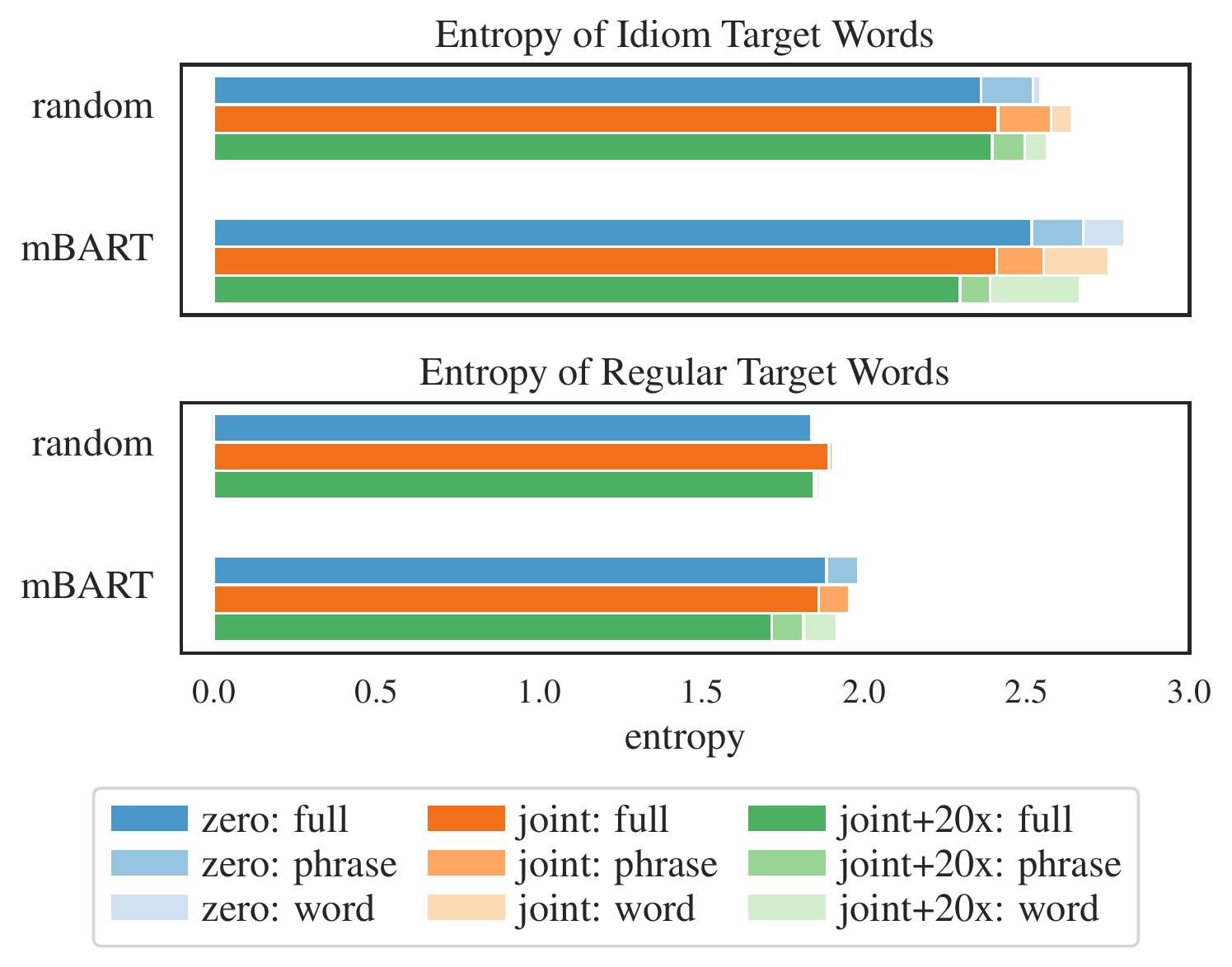}
    \caption{
    Model uncertainty for the translation of regular vs. idiom words.
    The length of each (lighter) bar encodes the difference from its (darker) bar to its left.
    }
    \label{fig:analysis-entropy}
\end{figure}

\subsection{Decoder Uncertainty}
\label{sec:analysis-entropy}

Next, we focus on how the token-level uncertainty of the decoder varies while it translates 
idiom vs. non-idiom words.
First, we translate each sentence pair with teacher-forcing and 
then measure the entropy of the decoder’s distributions for each (reference) target token.
Finally, using word alignments, we average separately the entropy values of target words\footnotemark~that are aligned to idiom and non-idiom source words.
\footnotetext{We map SPM tokens to words.}
Figure~\ref{fig:analysis-entropy} shows the results, 
in which \textit{lighter} shades correspond to \textit{narrower} contexts.

All models have significantly higher uncertainty 
when they translate idiom words (top section) compared to regular words (bottom section).
This confirms our expectation that it is harder to translate idioms.
When translating regular words,
the randomly initialized models are unaffected by changes in the idiom representations, 
whereas reducing the idiom context increases the uncertainty of the pretrained models.
This is another piece of evidence that pretraining yields less local models (\S\ref{sec:analysis-translation}).
When translating idiom words, 
including or upsampling the idiom-train data benefits the pretrained model,
but not the randomly initialized one.

\section{Discussion}

\paragraph{Global Metrics}
LiTER does not aim to replace global evaluation metrics like BLEU, but to complement them.
Global metrics estimate the general translation quality of model outputs, 
which is undoubtedly important.
However, they consider the full sentence, and as result the effects of idiom (mis)translation are overshadowed (see \S\ref{sec:results-regular-mt}).
LiTER aims to fill this gap by providing additional insights to practitioners with targeted evaluation.

\paragraph{LitTER}
LiTER should not be used in isolation, but combined with other (global/targeted) metrics.
The reason is that lower LiTER can be achieved by more accurate idiom translations or by hallucinations.
We aim to enable practitioners to make informed decisions without running
human evaluations in the model development phase.
The goal is to produce models that improve on LitTER without sacrificing general translation quality (e.g., BLEU).
Our experiments reveal this contrast, where unsupervised pretraining achieves this goal,
whereas (some of) the noisy variants fail.
We expect LiTER to be more relevant when developing NMT models for creative content (e.g., subtitles, social media text) that usually contains figurative language.

\paragraph{Alignment Metrics}
Word alignment-based methods aim to capture idiom translation accuracy that complements LitTER.
However, they are sensitive to the literal meaning of words to produce the alignments.
Thus, while alignment-based methods could be reliable in certain types of evaluation, such as gender translation~\cite{stanovsky-etal-2019-evaluating},
we believe that with the current techniques they should be used with caution for idiom translation evaluation. 
Although we did not systematically study this issue, we discovered by manual inspection that it was not uncommon to produce noisy or empty alignments.
We chose statistical over embedding-based methods as they yielded less empty alignments (\S\ref{sec:app-results-targeted-align}).

\section{Related Work}

\paragraph{Idioms in NMT}
There is limited research on idioms in NMT.
\citet{zaninello-birch-2020-multiword},
explore augmenting the training data with MWE translations from dictionaries, 
backtranslating~\cite{sennrich-etal-2016-improving} target-side sentences with MWEs, 
or wrapping the (source) MWEs with special tokens.
\citet{fadaee-etal-2018-examining} prepend a special token in source sentences that contain an idiom.
\citet{Gamallo2019-ti} do unsupervised translation of MWEs 
by composing cross-lingual word embeddings, 
but this is fundamentally incapable of translating idioms which are non-compositional.
Instead of using ad-hoc solutions that change the model or the data pipeline,
we use monolingual pretraining, which is a more generic and less invasive approach.

\paragraph{Targeted Evaluation}
Using word alignments is a straightforward approach
for the targeted evaluation of words or phrase translation~\cite{stanovsky-etal-2019-evaluating}.
\citet{fadaee-etal-2018-examining}, use word alignments to compare the reference and hypothesis matches that translate a given source idiom.
\citet{zaninello-birch-2020-multiword}
first align words in the hypothesis and the reference using edit-distance and then score the aligned words using character-level matching.
Alignment-based methods capture idiom translation accuracy,
but they do not account for literal translation errors
which are a major issue in idiom translation~\cite{fadaee-etal-2018-examining}.
The method of \citet{shao-etal-2018-evaluating} 
is capable of estimating the frequency of such errors, 
but it requires the creation of language-specific hand-crafted lists.
In this work, we lift this limitation
and enable the automatic evaluation of literal translation errors.

\section{Conclusions}
We present a comprehensive study of idiomatic expressions in NMT,
aiming to facilitate future research on the topic. 
We propose LitTER (\S\ref{sec:eval-litter}), a novel metric that enables the \textit{automatic} evaluation of literal translation errors.
LitTER is used for \textit{targeted} evaluation and aims to complement and not replace global metrics, such as BLUE or ChrF,
which consider the full sentence and can only measure the \textit{overall} translation quality.
We evaluate models in controlled conditions, with and without the test set idioms (i.e., zero-shot).
We explore pretraining on monolingual data for improving idiom translation,
as parallel idiom data is difficult to come by.
Interestingly, we find that pretraining achieves strong targeted improvements,
even in the zero-shot setting (\S\ref{sec:results-targeted}).
We also present a systematic analysis (\S\ref{sec:analysis}) that investigates the role of context in idiom translation. 
We find evidence that pretraining yields more contextual models, 
which helps to explain why it contributes to better idiom translations.
We also quantitatively confirm that idioms are more difficult to translate than regular words 
and strongly depend on the source context.

\section*{Limitations}

LitTER is a novel metric that sidesteps the need for human involvement to estimate the frequency of literal translation errors, which is a major problem in idiom translation by NMT models. However, in its current iteration it has certain limitations:

\begin{enumerate}
    \item An edge case with LitTER is that if the blocklist words appear as a result of translating other words - not part of the non-literal phrase - we will still count it as an error. We leave this as future work. 
    \item Our experiments were conducted on languages with relatively simple morphology.
    Preliminary experiments with German revealed that LitTER struggles with compound words.
    We did address these issues with custom rules, 
    but we aim to systematically study these cases in future versions of LitTER.
    \item The metric can be ambiguous when used in isolation. We recommend pairing it with standard evaluation metrics when comparing translation models.
\end{enumerate}

\bibliographystyle{acl_natbib}
\bibliography{anthology,acl2021}

\clearpage

\appendix

\section{Idiom Data Collection}
\label{sec:app-idiom-data}

\begin{figure}[t]
    \centering
    \includegraphics[width=1\columnwidth]{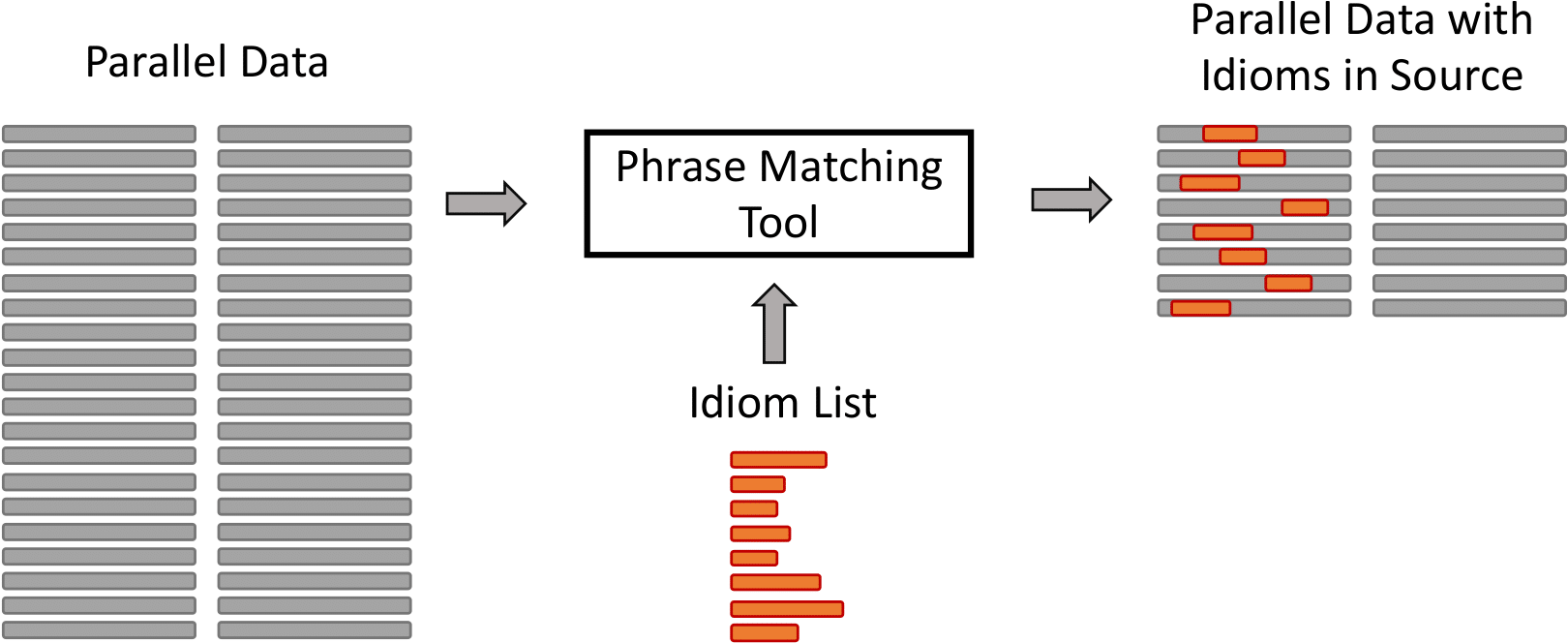}
    \caption{Overview of the process for collecting the idiom parallel data.}
    \label{fig:phrase-matching-process}
\end{figure}

We collected parallel data with idioms in the source side and 
annotated the spans in which the idioms occur within each sentence.
This enables us to conduct controlled experiments and to support our targeted evaluation metrics and our analysis.
To collect the data, 
we created a phrase-matching tool that searches for idioms in parallel data
and extracts and annotates the retrieved pairs (Figure~\ref{fig:phrase-matching-process}). 

\begin{figure}[t]
    \centering
    \fbox{
    \includegraphics[width=0.95\columnwidth]{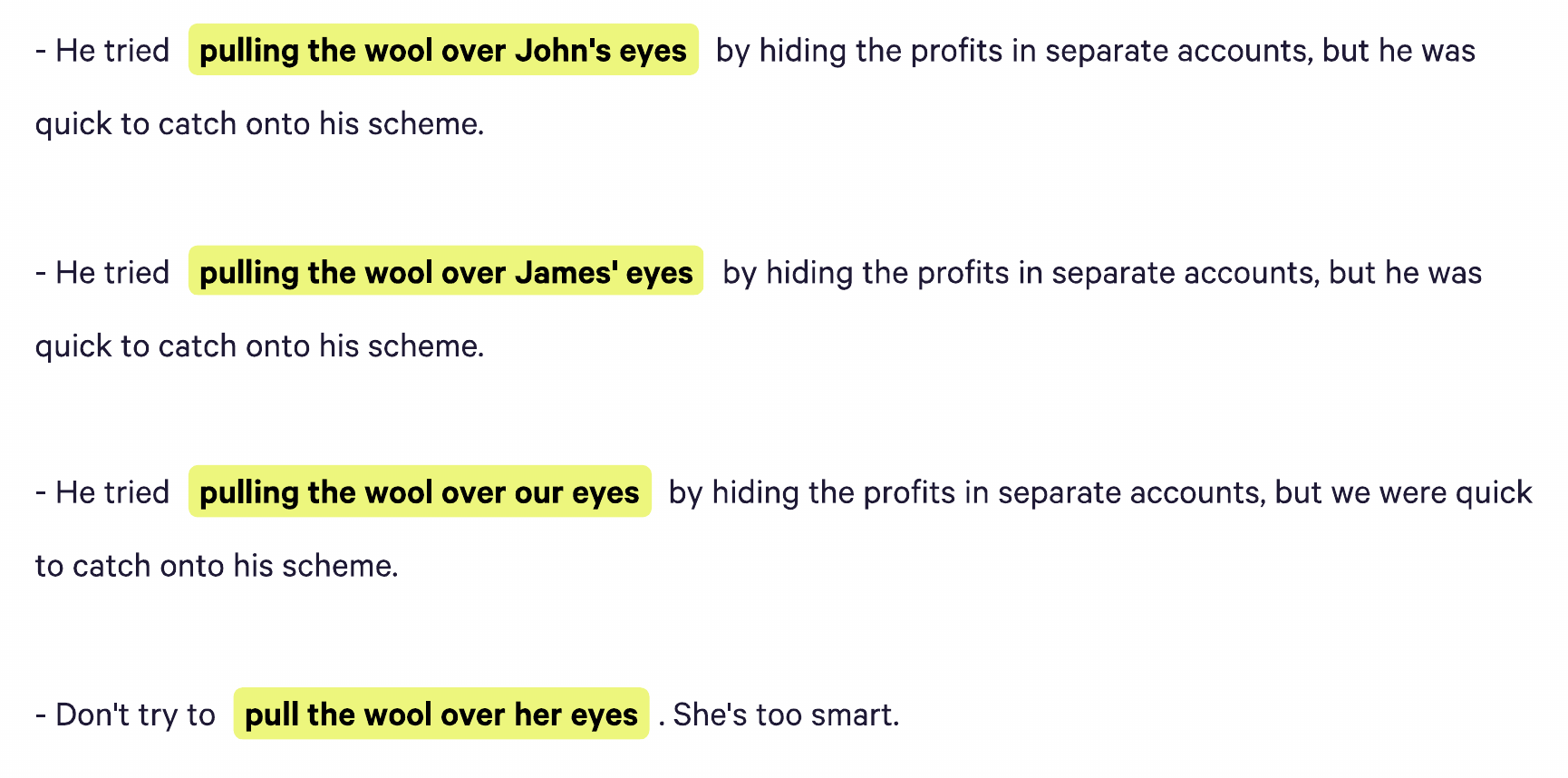}}
    \caption{Overview of the process for collecting the idiom parallel data.}
    \label{fig:example-phrase-matches}
\end{figure}

\paragraph{Phrase-Matching Tool}
Our tool, which we make publicly available, 
uses rule-based matching to search for sentences that contain phrases 
specified in user-defined phrase list.
While in this work we use it to create parallel data with idioms, 
it could be used to create datasets with different types of phrases.
We build our tool on top of Spacy's~\cite{spacy2} rule-based matching engine\footnotemark~that 
is more flexible and easy to work with than using regular expressions or custom rules~\cite{fadaee-etal-2018-examining}.
\footnotetext{\url{spacy.io/usage/rule-based-matching}}
It allows us to do pattern matching over linguistic units, 
such as parts-of-speech or even dependency relations, 
thus capturing complex variations of a given phrase. 
Our tool first reads the input phrase list, and for each phrase,
it automatically creates a pattern based on some simple rules and assumptions.
We created separate idiom train and test data for each language-pair 
and we describe the process in Section~\ref{sec:experiments-data}.

Figure~\ref{fig:example-phrase-matches}, shows an actual example\footnotemark~of the different variations of the idiom 
“pull the wool over someone's eyes” that our tool captures. Notice that it matches:
\begin{itemize}
    \item Different variations of the verb ``pull''.
    \item Different words in the place of the word ``someone''. 
    \item Importantly, it optionally matches the particle ``'s''. This shows that we can apply logic base on the part-of-speech (POS) or other linguistic properties of words, and in this case optionally skip them.
\end{itemize}

\footnotetext{You can view this example in this online \href{https://explosion.ai/demos/matcher?text=\%20-\%20He\%20tried\%20pulling\%20the\%20wool\%20over\%20John\%27s\%20eyes\%20by\%20hiding\%20the\%20profits\%20in\%20separate\%20accounts\%2C\%20but\%20he\%20was\%20quick\%20to\%20catch\%20onto\%20his\%20scheme.\%0A\%0A\%20-\%20He\%20tried\%20pulling\%20the\%20wool\%20over\%20James\%27\%20eyes\%20by\%20hiding\%20the\%20profits\%20in\%20separate\%20accounts\%2C\%20but\%20he\%20was\%20quick\%20to\%20catch\%20onto\%20his\%20scheme.\%0A\%0A\%20-\%20He\%20tried\%20pulling\%20the\%20wool\%20over\%20our\%20eyes\%20by\%20hiding\%20the\%20profits\%20in\%20separate\%20accounts\%2C\%20but\%20we\%20were\%20quick\%20to\%20catch\%20onto\%20his\%20scheme.\%0A\%0A\%20-\%20Don\%27t\%20try\%20to\%20pull\%20the\%20wool\%20over\%20her\%20eyes.\%20She\%27s\%20too\%20smart.\%0A&model=en_core_web_sm&pattern=\%5B\%7B\%22id\%22\%3A8\%2C\%22attrs\%22\%3A\%5B\%7B\%22name\%22\%3A\%22LEMMA\%22\%2C\%22value\%22\%3A\%22pull\%22\%7D\%5D\%7D\%2C\%7B\%22id\%22\%3A9\%2C\%22attrs\%22\%3A\%5B\%7B\%22name\%22\%3A\%22LEMMA\%22\%2C\%22value\%22\%3A\%22the\%22\%7D\%5D\%7D\%2C\%7B\%22id\%22\%3A10\%2C\%22attrs\%22\%3A\%5B\%7B\%22name\%22\%3A\%22LEMMA\%22\%2C\%22value\%22\%3A\%22wool\%22\%7D\%5D\%7D\%2C\%7B\%22id\%22\%3A11\%2C\%22attrs\%22\%3A\%5B\%7B\%22name\%22\%3A\%22LEMMA\%22\%2C\%22value\%22\%3A\%22over\%22\%7D\%5D\%7D\%2C\%7B\%22id\%22\%3A12\%2C\%22attrs\%22\%3A\%5B\%7B\%22name\%22\%3A\%22IS_ALPHA\%22\%2C\%22value\%22\%3Atrue\%7D\%2C\%7B\%22name\%22\%3A\%22OP\%22\%2C\%22value\%22\%3A\%22\%3F\%22\%7D\%5D\%7D\%2C\%7B\%22id\%22\%3A5\%2C\%22attrs\%22\%3A\%5B\%7B\%22name\%22\%3A\%22POS\%22\%2C\%22value\%22\%3A\%22PART\%22\%7D\%2C\%7B\%22name\%22\%3A\%22OP\%22\%2C\%22value\%22\%3A\%22\%3F\%22\%7D\%5D\%7D\%2C\%7B\%22id\%22\%3A6\%2C\%22attrs\%22\%3A\%5B\%7B\%22name\%22\%3A\%22ORTH\%22\%2C\%22value\%22\%3A\%22eyes\%22\%7D\%5D\%7D\%5D}{demo}. It shows the rule we generate for the idiom phrase and the variants it captures.}

\begin{figure}[t]
    \centering
    \includegraphics[width=1\columnwidth]{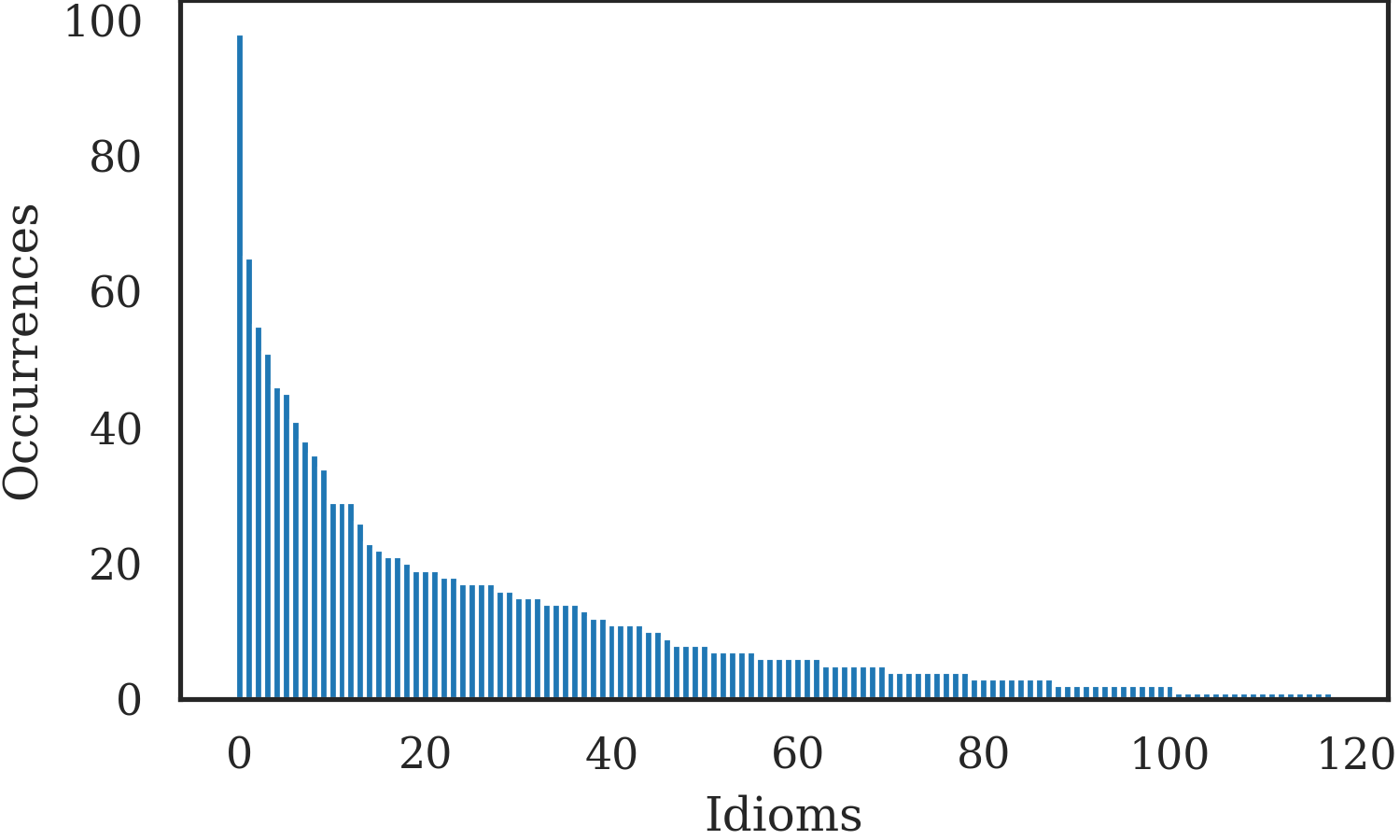}
    \caption{Occurrences of idioms in the en$\rightarrow$fr idiom-test sets.}
    \label{fig:app-test-idiom-stats-fr}
\end{figure}

\begin{figure}[t]
    \centering
    \includegraphics[width=1\columnwidth]{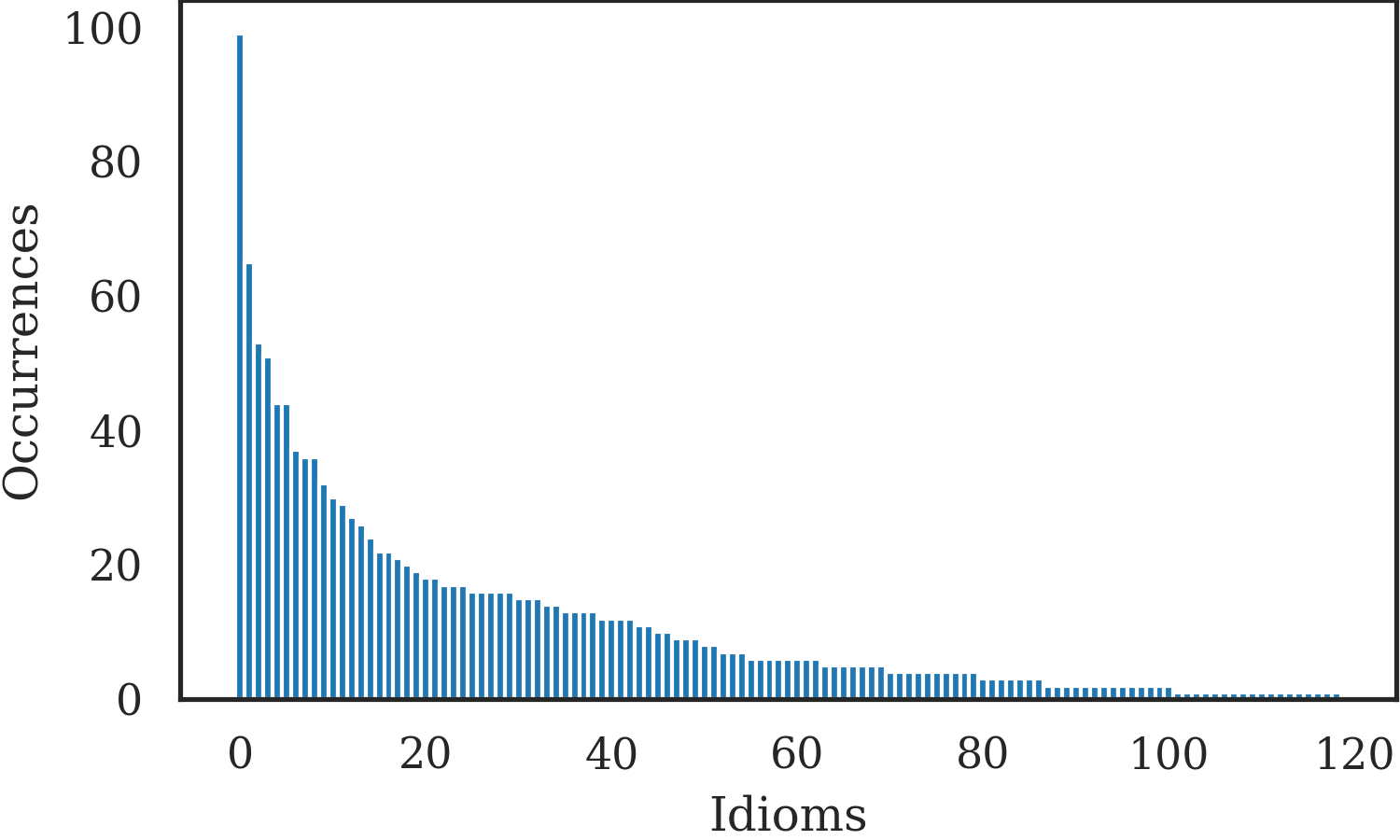}
    \caption{Occurrences of idioms in the en$\rightarrow$es idiom-test sets.}
    \label{fig:app-test-idiom-stats-es}
\end{figure}

\subsection{Idiom-test Statistics}
\label{sec:app-idiom-stats}
Figure~\ref{fig:app-test-idiom-stats-fr} and Figure~\ref{fig:app-test-idiom-stats-es}
show the occurrences of idioms in the en$\rightarrow$fr and en$\rightarrow$es idiom-test sets, respectively. We see that in both test sets the idiom statistics follow very similar distributions. This verifies that different idioms have very different frequencies, and highlights the need for macro-averaging the scores of targeted evaluation metrics.
Failure to do so would promote models that have perform best on the most frequent idioms over those that have a more balanced performance.

Also, note that the frequencies of idioms in the idiom-train and idiom-test sets are identical, as described in Section~\ref{sec:experiments-data}.
While we have not computed the idiom frequencies in the monolingual data used to pretrain mBART, we expect that they would follow a similar distribution as the one found in the monolingual data we used in our work.

\begingroup
\setlength{\tabcolsep}{5.5pt} %
\renewcommand{\arraystretch}{1} %
\begin{table*}[t]
    \tiny
    \centering
    \begin{tabular}{@{}clrrrrrrrrrrr@{}}
	\toprule[1.5pt]
	\multirow{2}{*}{Split}
	& \multirow{2}{*}{Model (en$\rightarrow$fr)} & \multirow{2}{*}{LitTER↓}
	& \multicolumn{2}{c}{APT-Eval (fast-align)}
	& \multicolumn{2}{c}{APT-Eval (awesome-align)}
	& \multicolumn{3}{c}{Global Evaluation BLEU↑}
	& \multicolumn{3}{c}{Global Evaluation ChrF↑}
	\\  \cmidrule(lr){4-5} \cmidrule(lr){6-7} \cmidrule(lr){8-10} \cmidrule(lr){11-13}
	                                                                          &                      &       & UniPrec↑   & ChrF↑   & UniPrec↑   & ChrF↑   & IWSLT17 & WMT14  & idiom & IWSLT17 & WMT14 & idiom \\\midrule
	\multirow{5}{*}{zero}                                                     & random               & 0.563 & 0.268      & 0.298   & 0.272      & 0.319   & 44.1    & 34.8   & 34.4  & 0.67    & 0.61  & 0.60  \\
	                                                                          & mBART                & 0.484 & 0.291      & 0.322   & 0.296      & 0.347   & 47.0    & 38.6   & 36.5  & 0.69    & 0.64  & 0.61  \\
	                                                                          & mBART +mask          & 0.478 & 0.298      & 0.323   & 0.294      & 0.339   & 46.3    & 38.2   & 36.0  & 0.68    & 0.64  & 0.61  \\
	                                                                          & mBART +replace (dec) & 0.519 & 0.295      & 0.319   & 0.304      & 0.346   & 46.9    & 39.0   & 36.0  & 0.69    & 0.64  & 0.61  \\
	                                                                          & mBART +replace (enc) & 0.365 & 0.260      & 0.284   & 0.262      & 0.306   & 44.1    & 36.2   & 34.5  & 0.66    & 0.62  & 0.60  \\\midrule
	\multirow{5}{*}{joint}                                                    & random               & 0.448 & 0.317      & 0.337   & 0.322      & 0.365   & 44.2    & 34.8   & 35.3  & 0.67    & 0.61  & 0.61  \\
	                                                                          & mBART                & 0.408 & 0.333      & 0.352   & 0.343      & 0.384   & 46.5    & 38.5   & 37.3  & 0.68    & 0.64  & 0.62  \\
	                                                                          & mBART +mask          & 0.443 & 0.315      & 0.338   & 0.334      & 0.379   & 46.2    & 38.3   & 37.2  & 0.68    & 0.64  & 0.62  \\
	                                                                          & mBART +replace (dec) & 0.447 & 0.317      & 0.342   & 0.331      & 0.379   & 46.8    & 38.8   & 37.0  & 0.69    & 0.64  & 0.62  \\
	                                                                          & mBART +replace (enc) & 0.364 & 0.300      & 0.322   & 0.306      & 0.350   & 44.5    & 36.6   & 35.6  & 0.66    & 0.62  & 0.61  \\\midrule
	\multirow{2}{*}{\begin{tabular}[c]{@{}c@{}}upsample \\20x\end{tabular}}   & random (20x)         & 0.371 & 0.323      & 0.353   & 0.331      & 0.382   & 44.4    & 34.7   & 35.3  & 0.67    & 0.61  & 0.61  \\
	                                                                          & mBART (20x)          & 0.289 & 0.329      & 0.346   & 0.338      & 0.376   & 46.6    & 38.7   & 36.0  & 0.68    & 0.64  & 0.61  \\\midrule
	\multirow{2}{*}{\begin{tabular}[c]{@{}c@{}}upsample \\100x\end{tabular}}  & random (100x)        & 0.378 & 0.314      & 0.343   & 0.325      & 0.363   & 43.8    & 34.7   & 33.9  & 0.66    & 0.61  & 0.60  \\
	                                                                          & mBART (100x)         & 0.289 & 0.337      & 0.358   & 0.347      & 0.389   & 46.8    & 38.2   & 35.8  & 0.69    & 0.64  & 0.61  \\

	\bottomrule[1.5pt]
\end{tabular}

\caption{Translation results in en$\rightarrow$fr. The results involve a single run, but include mutliple test sets and are consistent across the board.}
\label{table:enfr-full}
\end{table*}
\endgroup

\begingroup
\setlength{\tabcolsep}{5.5pt} %
\renewcommand{\arraystretch}{1} %
\begin{table*}[t]
    \tiny
    \centering
    \begin{tabular}{@{}clrrrrrrrrrrr@{}}
        \toprule[1.5pt]
        \multirow{2}{*}{Split}
        & \multirow{2}{*}{Model (en$\rightarrow$es)} & \multirow{2}{*}{LitTER↓}
        & \multicolumn{2}{c}{APT-Eval (fast-align)}
        & \multicolumn{2}{c}{APT-Eval (awesome-align)}
        & \multicolumn{3}{c}{Global Evaluation BLEU↑}
        & \multicolumn{3}{c}{Global Evaluation ChrF↑} \\
        \cmidrule(lr){4-5} \cmidrule(lr){6-7} \cmidrule(lr){8-10} \cmidrule(lr){11-13}

                                                                                  &                      &       & UniPrec↑ & ChrF↑ & UniPrec↑ & ChrF↑ & IWSLT17 & WMT13 & idiom & IWSLT17 & WMT13 & idiom \\ \midrule
        \multirow{5}{*}{zero}                                                     & random               & 0.541 & 0.350    & 0.364 & 0.351    & 0.370 & 36.0    & 31.5  & 38.9  & 0.62    & 0.57  & 0.63 \\
                                                                                  & mBART                & 0.476 & 0.383    & 0.385 & 0.368    & 0.390 & 38.5    & 34.0  & 40.8  & 0.64    & 0.59  & 0.64 \\
                                                                                  & mBART +mask          & 0.481 & 0.369    & 0.378 & 0.354    & 0.380 & 38.8    & 34.0  & 40.5  & 0.64    & 0.59  & 0.64 \\
                                                                                  & mBART +replace (dec) & 0.508 & 0.388    & 0.389 & 0.384    & 0.401 & 38.6    & 34.4  & 40.4  & 0.64    & 0.59  & 0.64 \\
                                                                                  & mBART +replace (enc) & 0.389 & 0.334    & 0.345 & 0.323    & 0.351 & 37.0    & 32.1  & 39.0  & 0.62    & 0.57  & 0.62 \\ \midrule
        \multirow{5}{*}{joint}                                                    & random               & 0.468 & 0.385    & 0.395 & 0.382    & 0.397 & 35.9    & 31.8  & 39.7  & 0.62    & 0.57  & 0.64 \\
                                                                                  & mBART                & 0.412 & 0.399    & 0.406 & 0.393    & 0.419 & 38.8    & 33.9  & 41.9  & 0.64    & 0.59  & 0.65 \\
                                                                                  & mBART +mask          & 0.418 & 0.389    & 0.402 & 0.388    & 0.410 & 38.7    & 33.9  & 41.9  & 0.64    & 0.59  & 0.65 \\
                                                                                  & mBART +replace (dec) & 0.443 & 0.402    & 0.408 & 0.402    & 0.414 & 38.5    & 33.9  & 41.7  & 0.64    & 0.59  & 0.65 \\
                                                                                  & mBART +replace (enc) & 0.352 & 0.352    & 0.372 & 0.355    & 0.378 & 36.7    & 32.2  & 39.7  & 0.62    & 0.57  & 0.63 \\ \midrule
        \multirow{2}{*}{\begin{tabular}[c]{@{}c@{}} upsample\\ 20x \end{tabular}} & random (20x)         & 0.400 & 0.410    & 0.424 & 0.415    & 0.440 & 35.9    & 31.7  & 40.0  & 0.62    & 0.57  & 0.64 \\
                                                                                  & mBART (20x)          & 0.301 & 0.422    & 0.435 & 0.419    & 0.444 & 38.8    & 34.0  & 40.6  & 0.64    & 0.59  & 0.64 \\ \midrule
        \multirow{2}{*}{\begin{tabular}[c]{@{}c@{}} upsample\\ 100x \end{tabular}}& random (100x)        & 0.391 & 0.406    & 0.420 & 0.407    & 0.435 & 36.2    & 31.8  & 38.9  & 0.62    & 0.57  & 0.63 \\
                                                                                  & mBART (100x)         & 0.299 & 0.416    & 0.427 & 0.406    & 0.441 & 38.7    & 33.9  & 40.5  & 0.64    & 0.59  & 0.64 \\

        \bottomrule[1.5pt]
    \end{tabular}

    \caption{Translation results in en$\rightarrow$es. The results involve a single run, but are consistent across the board.}
    \label{table:enes-full}
\end{table*}
\endgroup

\FloatBarrier

\section{Translation Results}
\label{sec:app-results}

In this section, we present all of our translation results in detail.
Table~\ref{table:enfr-full}, shows our results in en$\rightarrow$fr,
while Table~\ref{table:enes-full} shows our results in en$\rightarrow$es.
Besides BLEU, we also include results with ChrF~\cite{popovic-2015-chrf}, 
for global translation evaluation.
overall the results are \textit{consistent} across language pairs in all evaluation methods.
The fact that the absolute scores reached by the model on each each test set are different is natural, as the test sets themselves are different to each other between languages.
However, the relative performance between model across language pairs is the same, with only minor differences.

\subsection{Targeted Evaluation}
\label{sec:app-results-targeted}
\begin{figure*}[t]
\centering
\begin{subfigure}{.44\textwidth}
  \centering
  \includegraphics[width=1\linewidth]{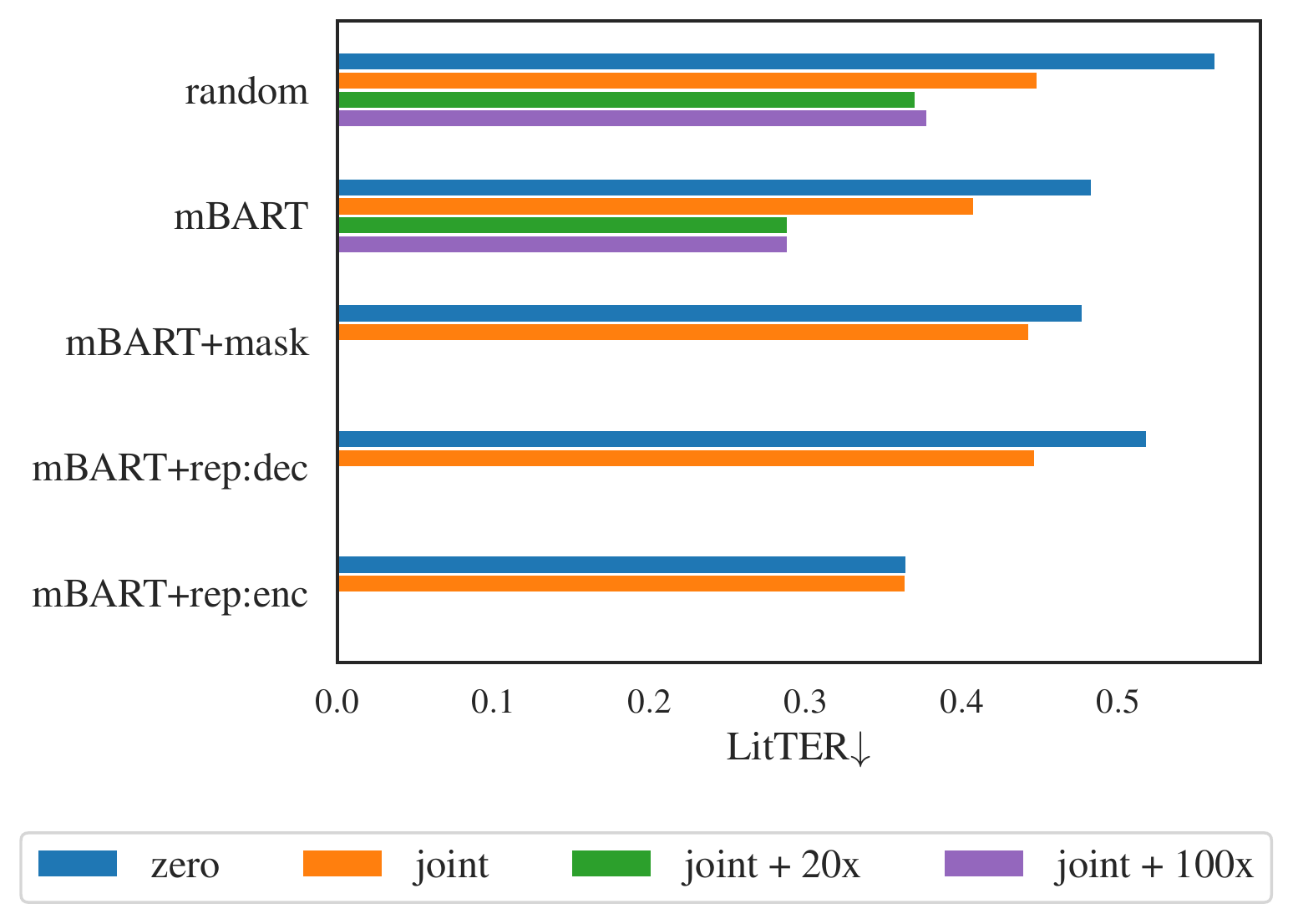}
  \caption{Results on LitTER for en$\rightarrow$fr.}
  \label{fig:app-enfr-litter}
\end{subfigure}%
\hfill
\begin{subfigure}{.44\textwidth}
  \centering
  \includegraphics[width=1\linewidth]{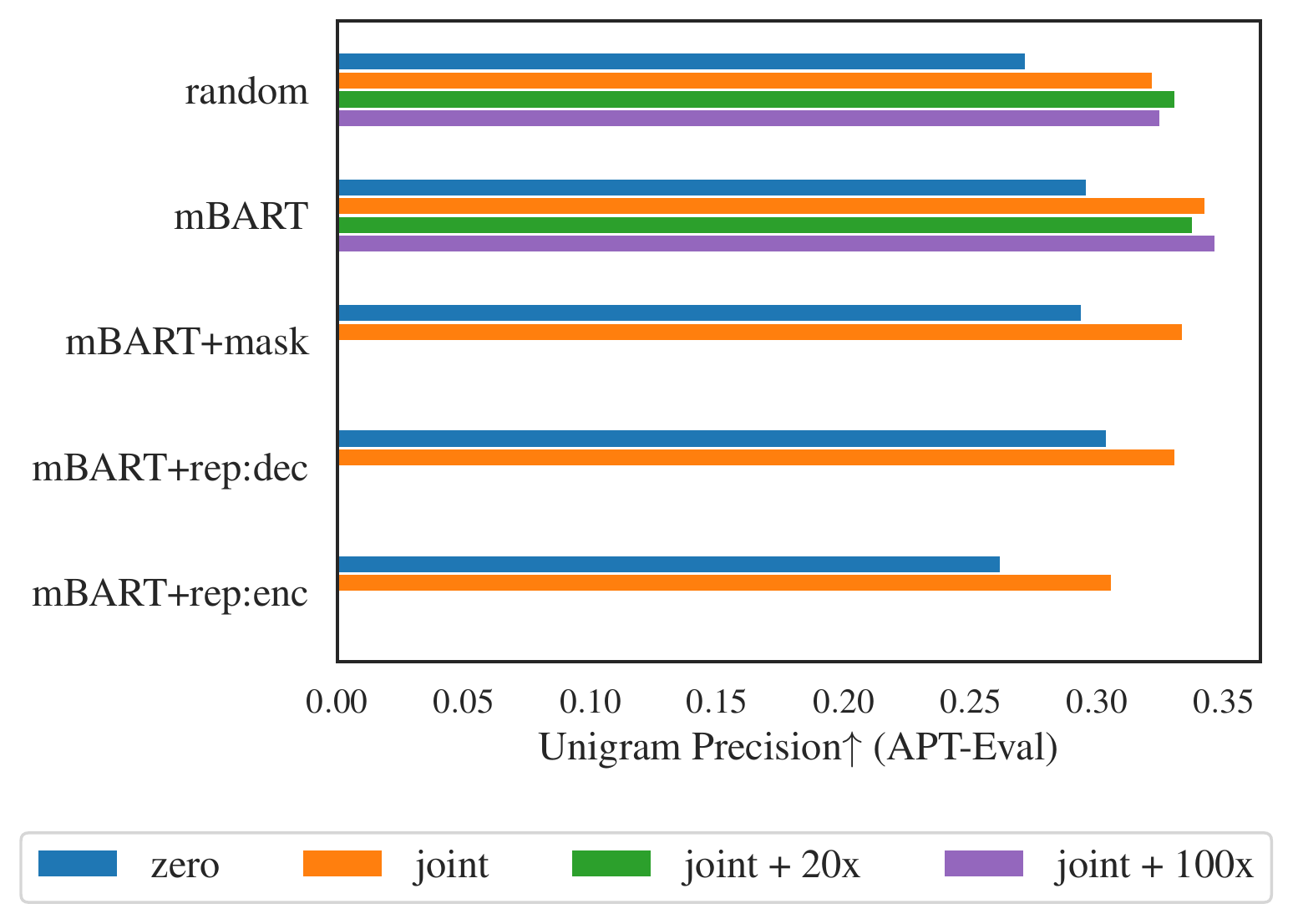}
  \caption{Results on APT-Eval (unigram precision) for en$\rightarrow$fr.}
  \label{fig:app-enfr-apt-uprec}
\end{subfigure}
\centering
\begin{subfigure}{.44\textwidth}
  \centering
  \includegraphics[width=1\linewidth]{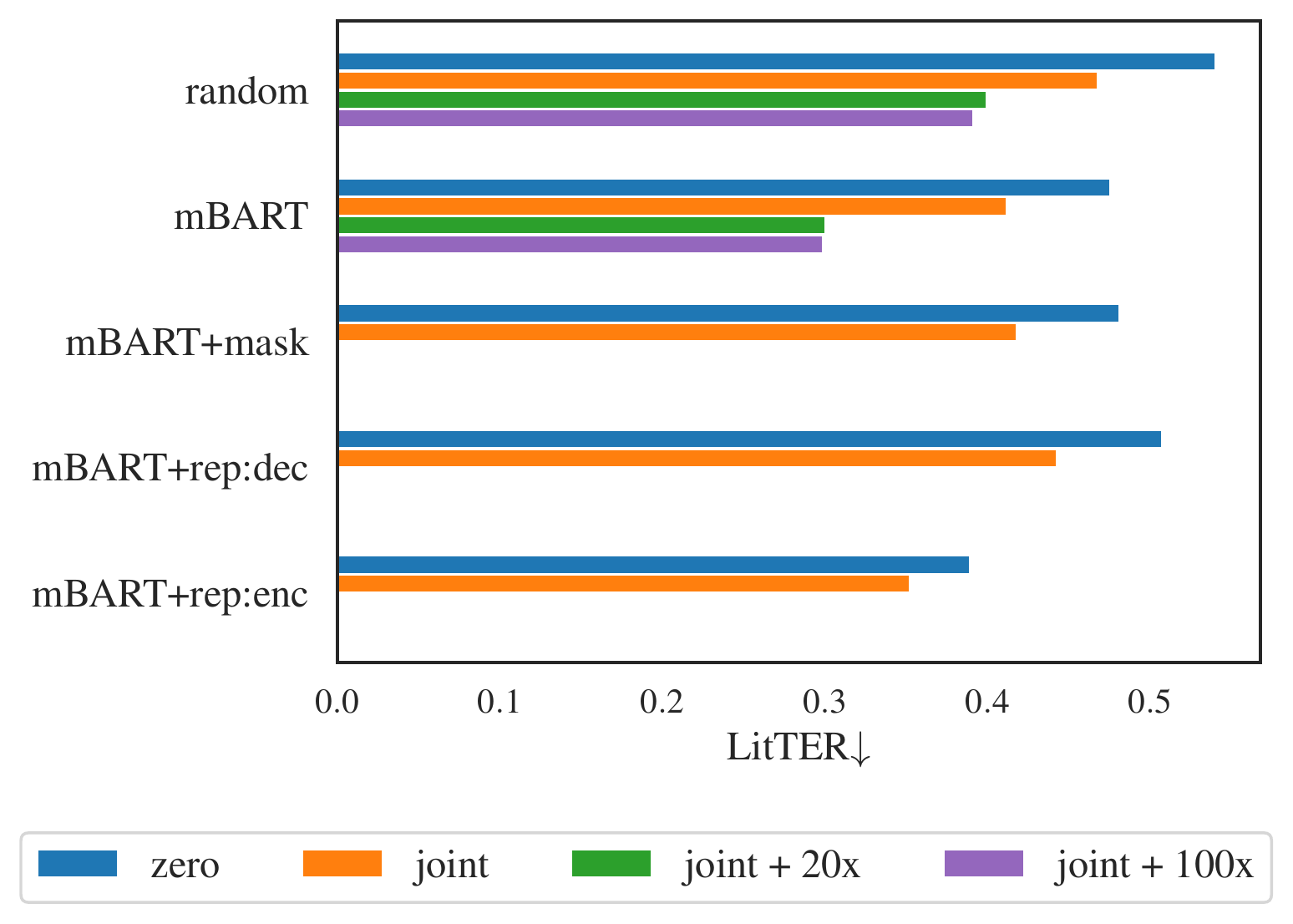}
  \caption{Results on LitTER for en$\rightarrow$es.}
  \label{fig:app-enes-litter}
\end{subfigure}%
\hfill
\begin{subfigure}{.44\textwidth}
  \centering
  \includegraphics[width=1\linewidth]{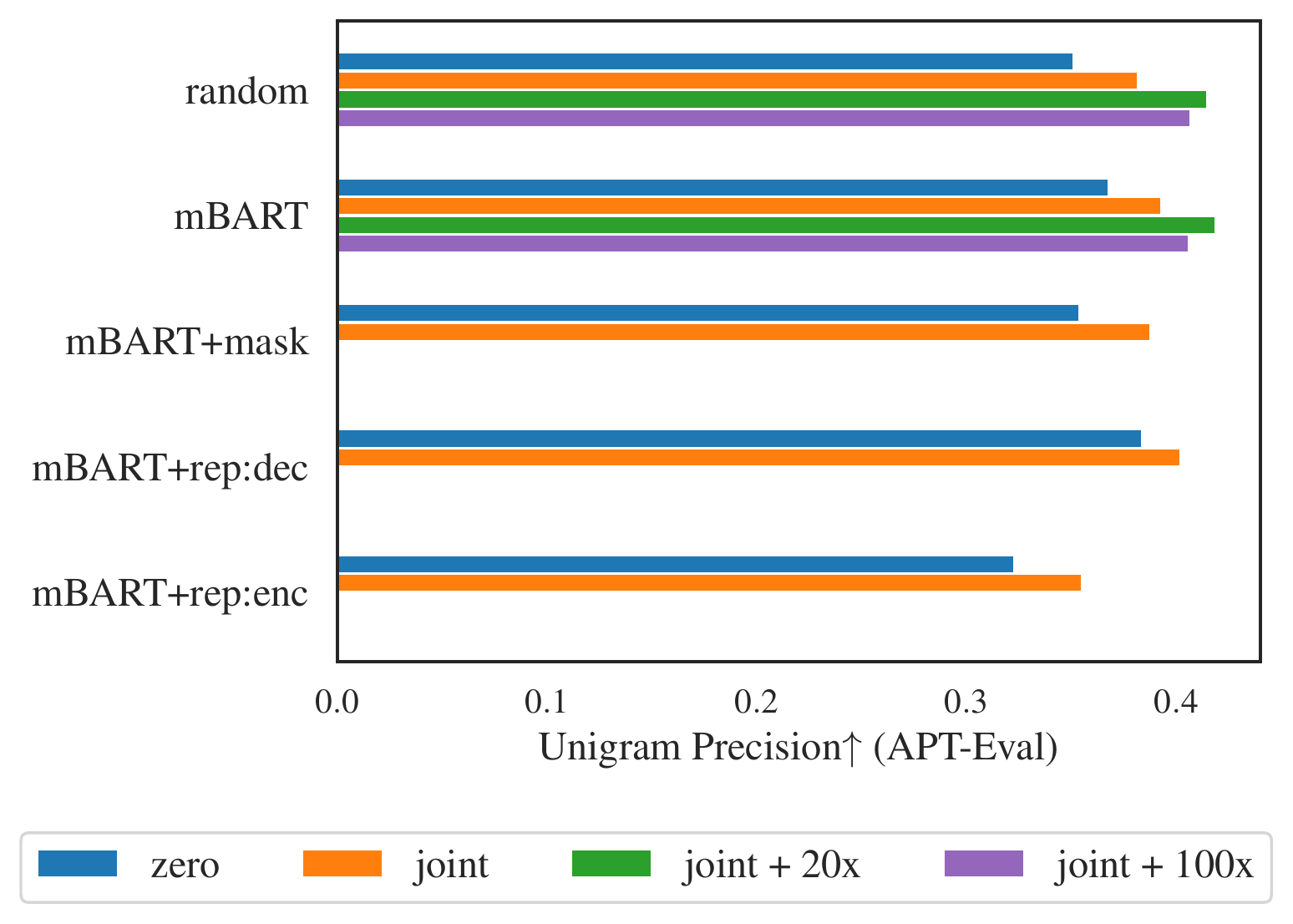}
  \caption{Results on APT-Eval (unigram precision) for en$\rightarrow$es.}
  \label{fig:app-enes-apt-uprec}
\end{subfigure}
\caption{Results on \textit{targeted} evaluation of idiom translation.
}
\label{fig:app-targeted}
\end{figure*}

In Figure~\ref{fig:app-targeted}, we visualize how model perform on our targeted evaluation methods for both en$\rightarrow$fr and en$\rightarrow$es.
Recall that, LitTER, measures how often each model makes literal translation errors.
APT-Eval, measures idiom translation accuracy, by comparing the reference and (aligned) hypothesis spans that translate the source idiom words.
In these plots, we use present how models performed on unigram precision metric for APT-Eval, similar to the main paper.
We observe that both in terms of literal translation errors (Figure~\ref{fig:app-enfr-litter},~\ref{fig:app-enes-litter}), 
as well as idiom translation accuracy (Figure~\ref{fig:app-enfr-apt-uprec},~\ref{fig:app-enes-apt-uprec}), 
the results are remarkably consistent across languages.
Upsampling the idiom-train data helps all models regardless of initialization, 
but upsampling more than 20x does not yield consistent improvements.

\subsection{Word Alignment Models}
\label{sec:app-results-targeted-align}

APT-Eval, requires word alignments to map the source idiom words to the words of the reference and hypothesis spans, respectively.
In our main paper, we presented results using fast-align~\cite{dyer-etal-2013-simple} with word alignment models trained on the training data of each language-pair.
We also experimented with awesome-align~\cite{dou-neubig-2021-word}, that doesn't require any training, 
and uses the token similarities of the pretrained mBERT models to obtain the alignments.
After analysis, we found that fast-align produced empty matches for the idiom words in 2.2\% of the reference sentences in our idiom-test set.
Awesome-align, however, yielded more empty matches, which after tweaking its threshold parameter\footnotemark, 
we managed to reduce it to 3\%.
While we omitted the awesome-align results from the main paper, 
we include them in Tables~\ref{table:enfr-full},~\ref{table:enes-full} for completeness.
We observe that while the absolute APT-Eval scores are different between the two alignment methods, 
the relative performance across models is consistent.

\footnotetext{We used the following hyper-parameters:\\\texttt{extraction=softmax, softmax\_threshold=0.001}}

\subsection{Regular Translation Evaluation}
\label{sec:app-results-regular-mt}
\begin{figure*}[t]
\centering
\begin{subfigure}{.44\textwidth}
  \centering
  \includegraphics[width=1\columnwidth]{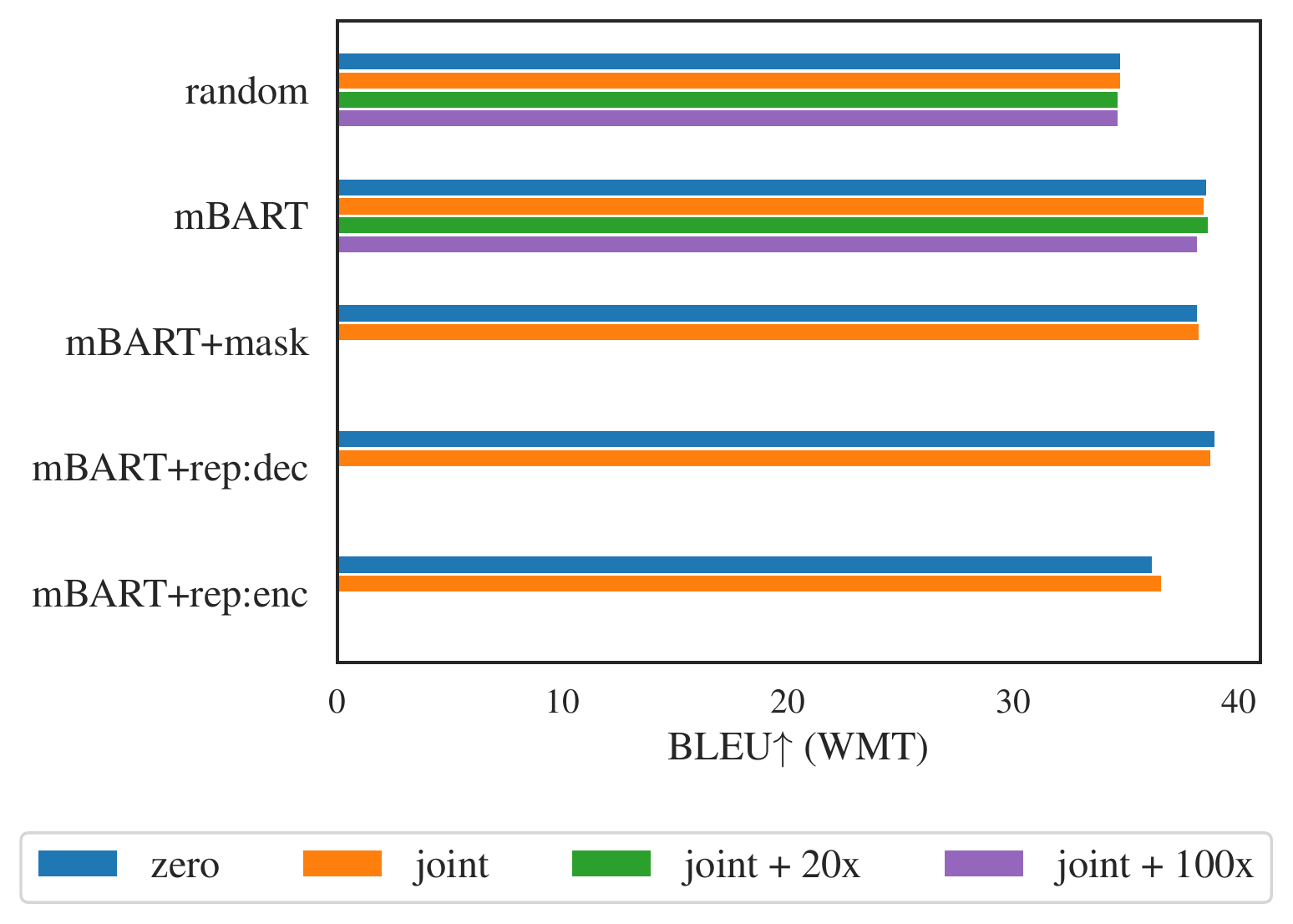}
  \caption{BLEU results on the en$\rightarrow$fr WMT14 test set}
  \label{fig:app-enfr-bleu-wmt}
\end{subfigure}%
\hfill
\begin{subfigure}{.44\textwidth}
  \centering
    \includegraphics[width=1\columnwidth]{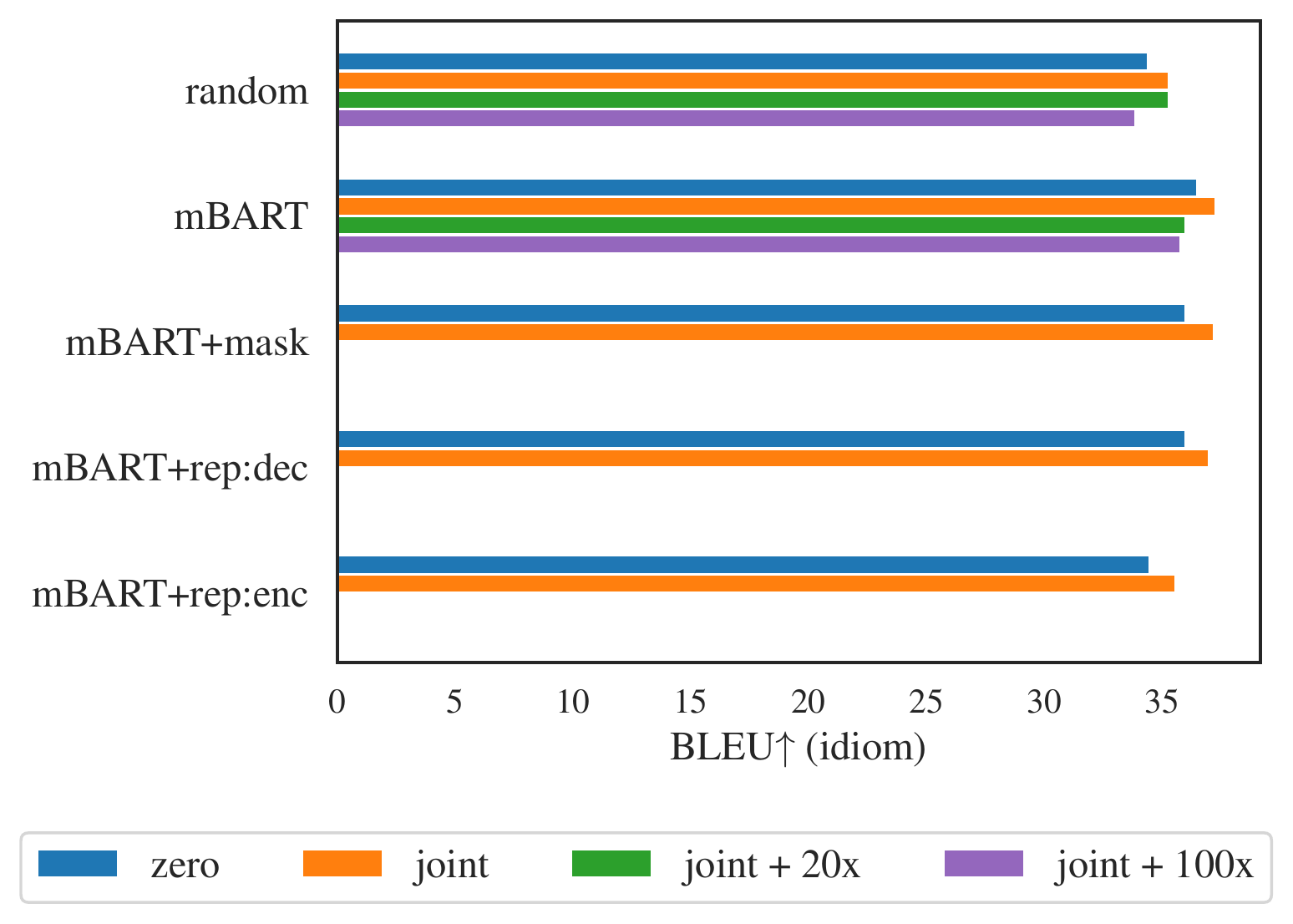}
  \caption{Regular BLEU results on our en$\rightarrow$fr \textit{idiom-test} set.}
  \label{fig:app-enfr-bleu-idiom}
\end{subfigure}
\begin{subfigure}{.44\textwidth}
  \centering
  \includegraphics[width=1\columnwidth]{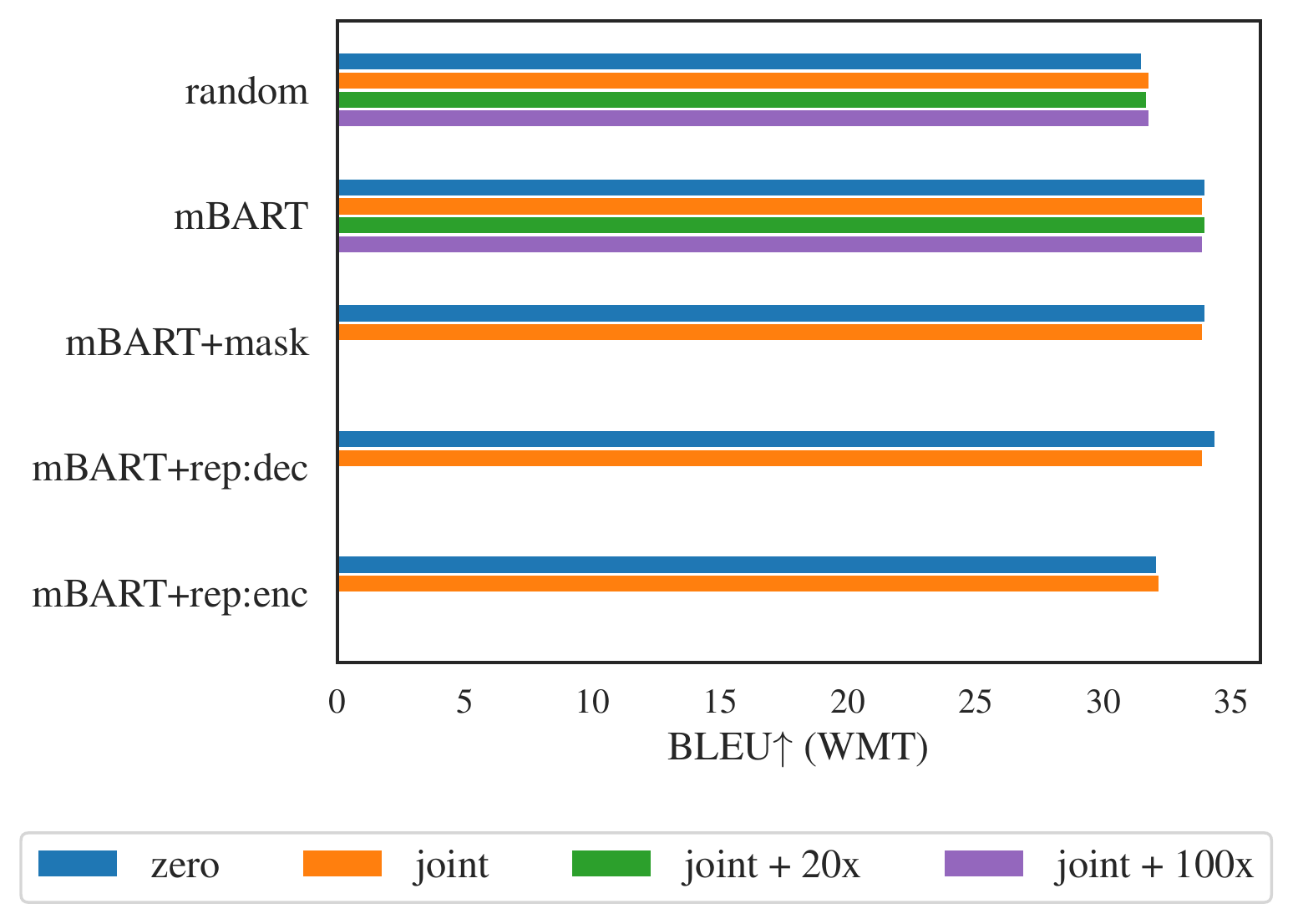}
  \caption{BLEU results on the en$\rightarrow$es WMT13 test set}
  \label{fig:app-enes-bleu-wmt}
\end{subfigure}%
\hfill
\begin{subfigure}{.44\textwidth}
  \centering
    \includegraphics[width=1\columnwidth]{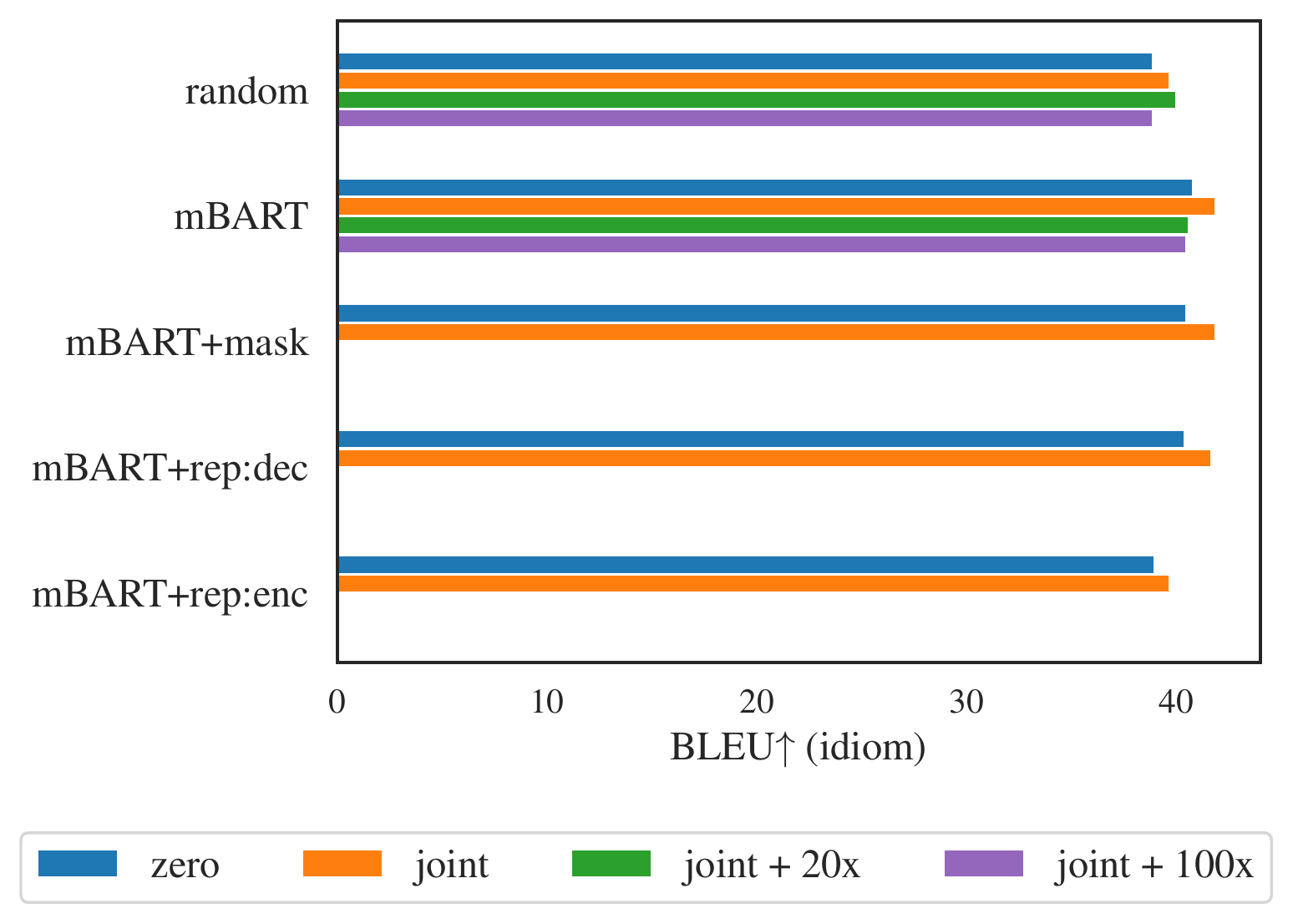}
  \caption{Regular BLEU results on our en$\rightarrow$es \textit{idiom-test} set.}
  \label{fig:app-enes-bleu-idiom}
\end{subfigure}
\caption{Results on \textit{regular} MT evaluation with BLEU on generic as well as on our idiom-test sets.}
\label{fig:app-app-generic}
\end{figure*}

Here, we visualize our results on \textit{regular} MT evaluation for both en$\rightarrow$fr and en$\rightarrow$es.
In Figure\ref{fig:app-app-generic}, we compare models in both language pairs and for both generic test sets as well as on our idiom-test sets.
Overall, we observe that the results are very consistent between language pairs, 
similar to the targeted evaluation results (\S\ref{sec:app-results-targeted}),
which improves our confidence in them.
As we already noted in our paper, we find that including or upsampling idiom training data has no measurable effect on generic test sets, unlike on our idiom-test set.

\newpage
\clearpage
\definecolor{green}{rgb}{0.0, 0.5, 0.0}

\FloatBarrier
\newpage
\clearpage

\section{LitTER}
\label{sec:app-litter}

\subsection{Indicative Examples}
\label{sec:app-litter-examples}

Here, we present some indicative examples of how LitTER evaluates translation outputs in our idiom test data. 
Figure~\ref{fig:litter-good-examples}, 
contains examples where LitTER is working as intended, whereas Figure~\ref{fig:litter-bad-examples}, contains a two failures of LitTER.

\newcommand{\samplesize}{0.45}

\begin{figure}[bht]
\centering

\small

\fbox{\begin{minipage}{\samplesize\textwidth}

\textbf{SRC:} As the example of Cyprus shows, Ankara does not \underline{pull its punches}. \\
\textbf{REF:} Comme le montre l'exemple de Chypre, Ankara n'y va pas avec le dos de la cuiller. \\
\textbf{HYP:} Comme le montre l'exemple de Chypre, Ankara ne tire pas les ficelles. \\

\textbf{Blocklists}\\
pull $\rightarrow$ \{tirez, tirer\}\\
its $\rightarrow$ \{ses, son, sa\}\\
punches $\rightarrow$ \{coups\}\\

No error detected.
\end{minipage}}

\smallskip

\fbox{\begin{minipage}{\samplesize\textwidth}

\textbf{SRC:} [..] it was already being \underline{put on ice} on the grounds that 'We'll never get it though the G20'.\\
\textbf{REF:} [..] elle était mise au rencart au motif que "nous n'arriverons jamais à convaincre le G20".\\
\textbf{HYO:} [..] on l'a déjà gelé au motif que "nous n'y arriverons jamais par le biais du G20".\\

\textbf{Blocklists}\\
put $\rightarrow$ \{mis, mettre\}\\
on $\rightarrow$ \{sur\}\\
ice $\rightarrow$ \{glace, ice, verglas\}\\

No error detected.
\end{minipage}}

\caption{LitTER failures on our en$\rightarrow$fr idiom test set. In the first example, the blocklist is not triggered because the inflected form \textit{tire} is missing from the blocklist. In the second example, the verb form \textit{gelé} (freeze) is not contained in the blocklist but is a literal translation of \textit{ice} in a wider sense.}
\label{fig:litter-bad-examples}

\end{figure}

\begin{figure}[t]
\centering
\small

\fbox{\begin{minipage}{\samplesize\textwidth}

\textbf{SRC:} To postpone this vote one more time would be to \underline{bark up the wrong tree}.\\
\textbf{REF:} Postposer ce vote une fois de plus eut été se tromper de cible.\\
\textbf{HYP:} Reporter ce vote une fois de plus, c'est se tromper d'arbre.\\

\textbf{Blocklists}\\
bark$\rightarrow$ \{aboyer, ecorces, ecorce\}\\
up$\rightarrow$  \{debout\}\\
the$\rightarrow$  \{le, la, les\}\\
wrong$\rightarrow$  \{faux, tort, errone, mal\}\\
tree$\rightarrow$  \{arbre, arbres, sapin, arborescence\}\\

\textbf{\color{red}{ERROR}:} Blocklist triggered by \{arbre\}

\end{minipage}}

\smallskip

\fbox{\begin{minipage}{\samplesize\textwidth}

\textbf{SRC:} For companies, using technology to gather important data, its like \underline{bread and butter}.\\
\textbf{REF:} Pour les sociétés, utiliser la technologie pour recueillir des données, c'est la routine.\\
\textbf{HYP:} Pour les entreprises, utiliser la technologie pour collecter des données importantes, c'est comme du pain et du beurre.\\

\textbf{Blocklists}\\
bread$\rightarrow$ \{pain\}\\
and$\rightarrow$  \{et\}\\
butter$\rightarrow$  \{et, pain, beurre\}\\

\textbf{\color{red}{ERROR}:} Blocklist triggered by \{et, pain, beurre\}

\end{minipage}}

\smallskip

\fbox{\begin{minipage}{\samplesize\textwidth}

\textbf{SRC:} And here is some \underline{eye candy} for you, from a range of DIY scientists and artists from all over the globe.\\
\textbf{REF:} Et voici quelques bonbons pour vos yeux, de la part d'un éventail de scientifiques et des artistes bricoleurs de tous les coins de la planète.\\
\textbf{HYP:} Et voici quelques bonbons pour les yeux, d'une gamme de scientifiques et d'artistes du bricolage du monde entier.\\

\textbf{Blocklists}\\
eye $\rightarrow$ \{oculaire, oeil, yeux, œil\}\\
candy $\rightarrow$ \{bonbon, bonbons, sucrerie\}\\

No error detected.

\end{minipage}}

\caption{Examples of LiTER evaluation on sentences in our en$\rightarrow$fr idiom test set. In the first two examples, the model makes a literal translation error and the error is captured by LitTER. In the third example, the literal translation is correct and the blocklist is not triggered, thanks to the 3rd step in our algorithm (\S~\ref{sec:eval-litter}). 
}
\label{fig:litter-good-examples}
\vspace{4cm}

\end{figure}

\clearpage
\newpage

\section{Analysis}
\label{sec:app-analysis}

In this section we present all of our analysis results.
Specifically, we include results on en$\rightarrow$es, with the (mBART) pretrained models finetuned with different noising methods, and additional probes.
For the noisy versions of mBART finetuning, we present results on the zero and joint split.
Recall that, in most of our probes, we evaluate the role of (idiom) context idiom translation. 
To do this,  we encode the idiom words within different contexts
and compare how this affects various aspects of each model.
Figure \ref{fig:analysis-overview}, illustrates the process by which we obtain the encoding for each context. We consider the following context:
(1) \textit{full context}, 
in which we encode the idiom phrase within the whole input sentence,
(2) \textit{phrase-level context}, 
in which we encode together only the words idiom phrase,
(3) \textit{word-level context}, 
in which we encode each idiom word independently.

\begin{figure*}[t]
\centering
\begin{subfigure}{.48\textwidth}
  \centering
  \includegraphics[width=1\columnwidth]{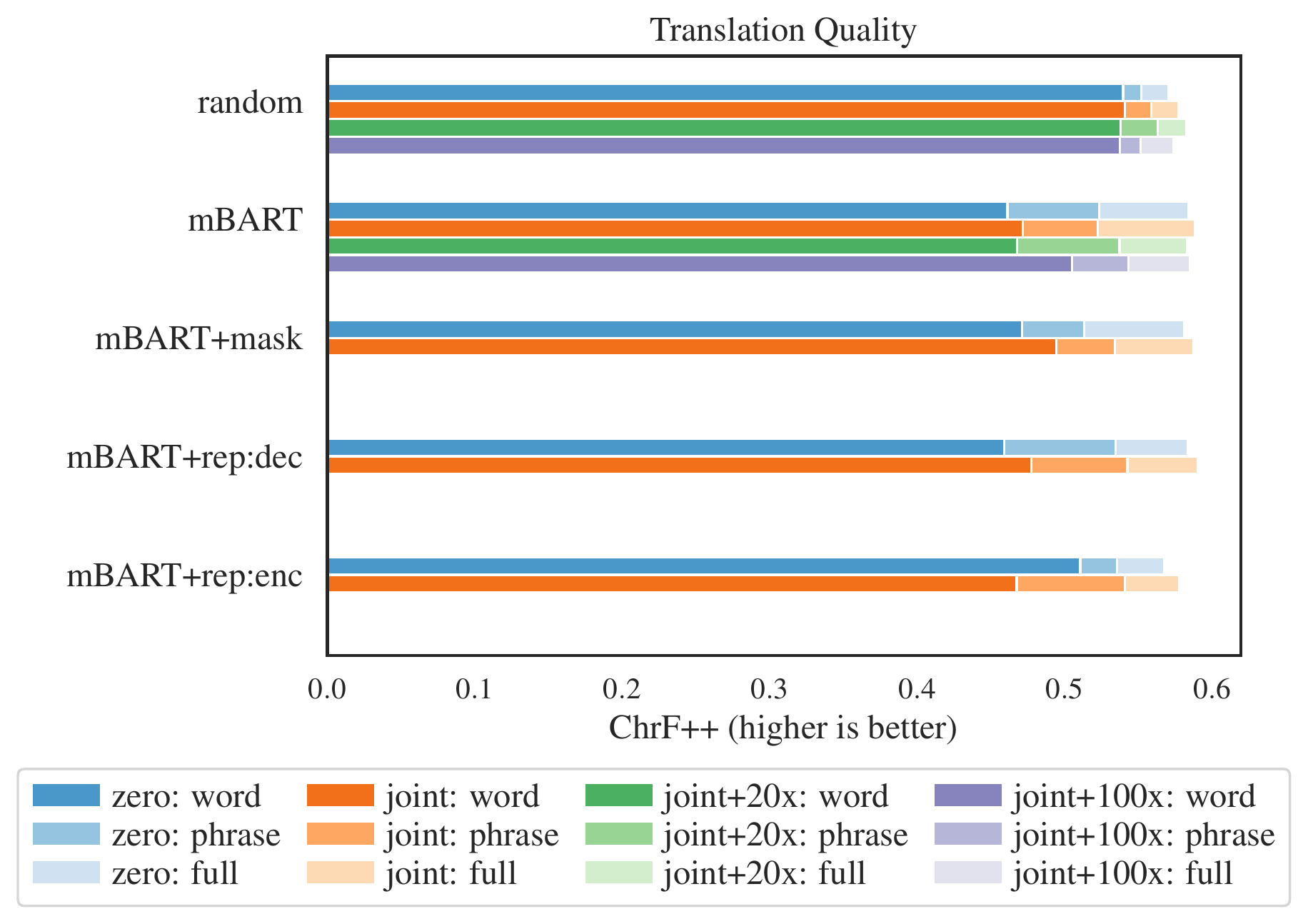}
  \caption{Results for the en$\rightarrow$fr models.}
  \label{fig:app-analysis-mt-enfr}
\end{subfigure}%
\hfill
\begin{subfigure}{.48\textwidth}
  \centering
  \includegraphics[width=1\columnwidth]{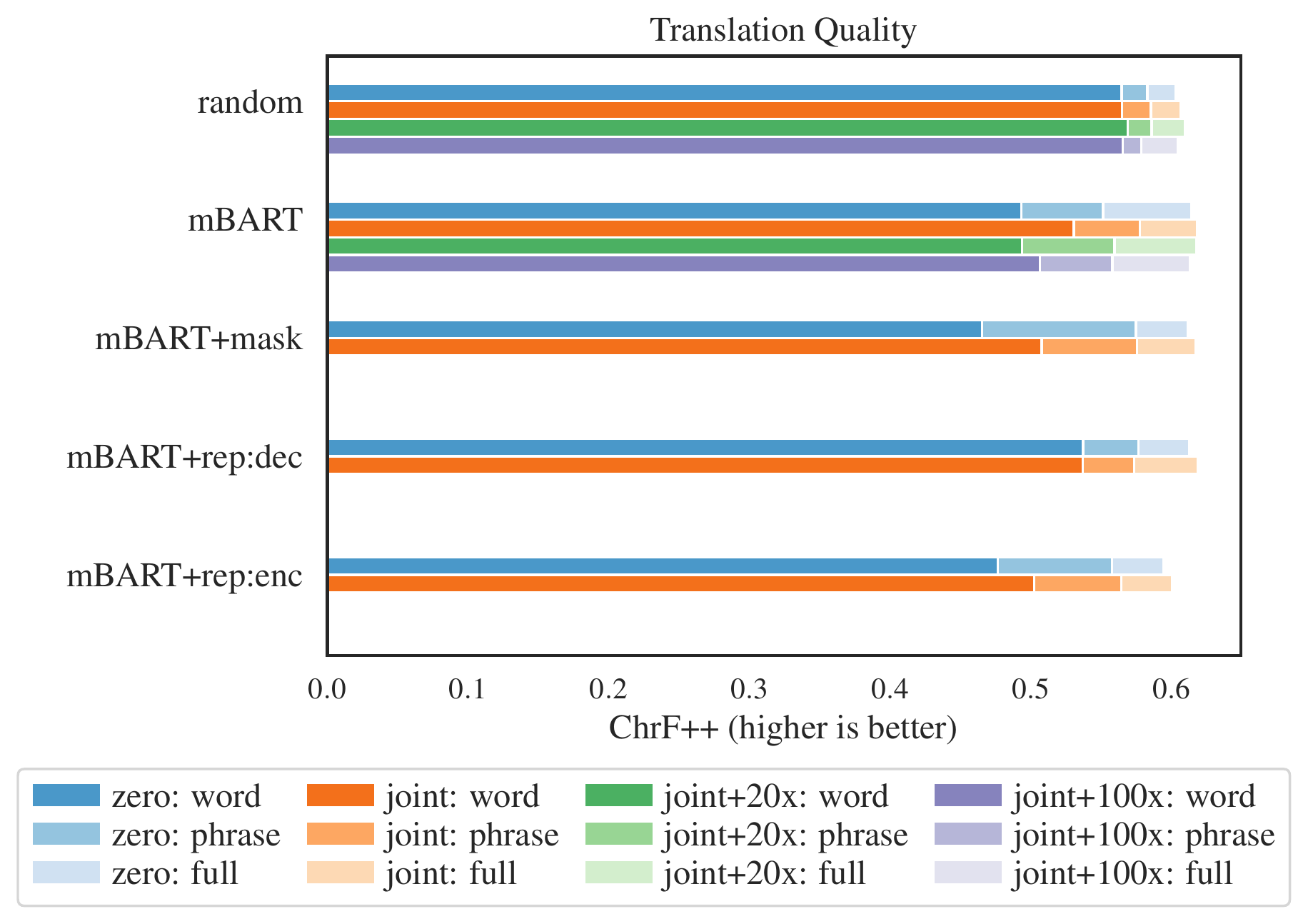}
  \caption{Results for the en$\rightarrow$es models.}
  \label{fig:app-analysis-mt-enes}
\end{subfigure}
    \caption{
    Variation in translation quality, measured in ChrF, as we vary the idiom representations.
    The length of each (lighter) bar encodes the difference from its (darker) bar to its left
    (i.e., overlapping bars effect).
    The darkest shades correspond to using the full (original) context for encoding idiom words, and each lighter shades correspond to narrower contexts.
    Overall, the results are consistent, with the exception of the ``mBART+replace(enc)'' model.}
\label{fig:app-analysis-mt}
\end{figure*}

\begin{figure}[t]
\centering
\includegraphics[width=1\columnwidth]{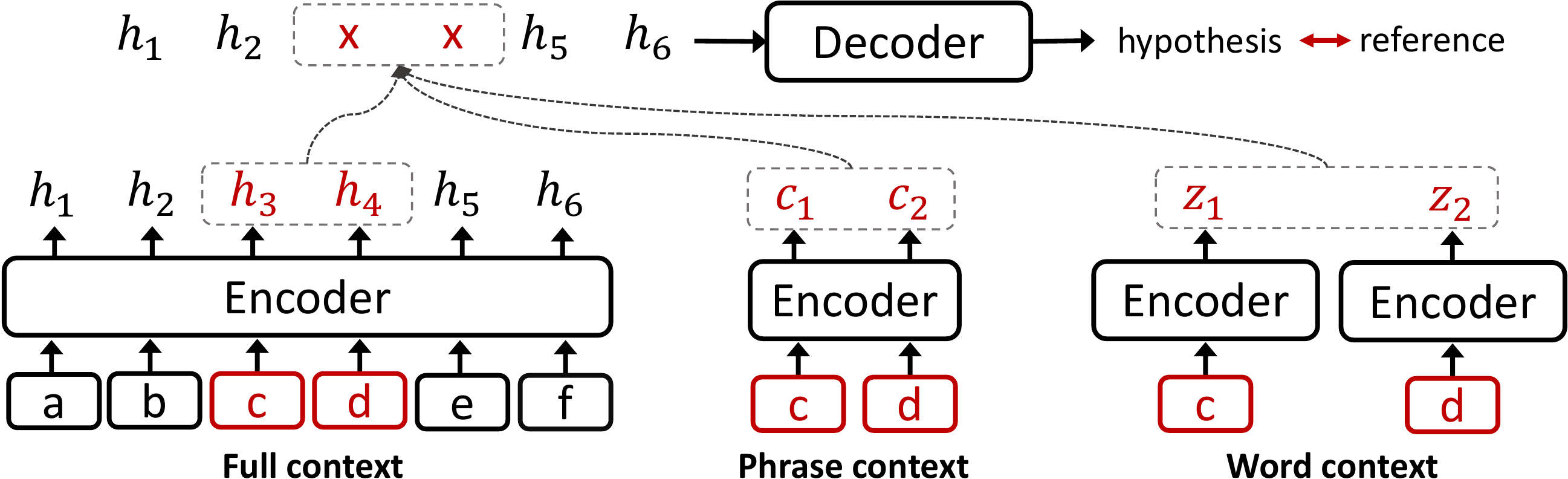}    
\caption{Illustration of how we probe the impact of different idiom contexts on translation quality.
First, we obtain each sequence of idiom token representations, by encoding them within different contexts.
Then we feed each sentence to the encoder, and before passing its outputs to the decoder, 
we replace the idiom token representations with those whose context we want to probe.
Finally, we sample a translation from the decoder and compare it against the reference.}
\label{fig:app-analysis-translation}
\end{figure}

\subsection{Variation in Translation Performance}
\label{sec:app-analysis-translation}

With this probe, we test how the variation of the idiom encoder representations is reflected in the translation output.
Figure~\ref{fig:app-analysis-translation}, shows a visual example of how this probe works.
First we encode each input sentence 
and then replace the encoder output representations belonging only to idiom words 
with those obtained with different (narrower) contexts.
Finally, we decode each encoder output sequence and compare the generated translation to the reference translation.

We observe that when using the full context, the results are generally consistent across language pairs and models.
As we discussed in the main paper, the mBART-initialized model suffers greatly when the idiom representations are encoded with narrower context, in contrast to the randomly initialized model. 
We believe this is an indication that the mBART-initialized model is less local, meaning that each token representation contains to a large degree information about the rest of the tokens.
We also observe that this behaviour is exhibited by all pretrained models.

However, while results are generally consistent in both language pairs, 
we do see some \textit{small} discrepancies as we probe the effects of narrower contexts,
in particular for the ``mBART+replace(enc)'' model.
We do not have a satisfying explanation for this performance difference, which occurs only when we encode idioms with word context.

\begin{figure*}[t]
\centering
\begin{subfigure}{.48\textwidth}
  \centering
  \includegraphics[width=1\columnwidth]{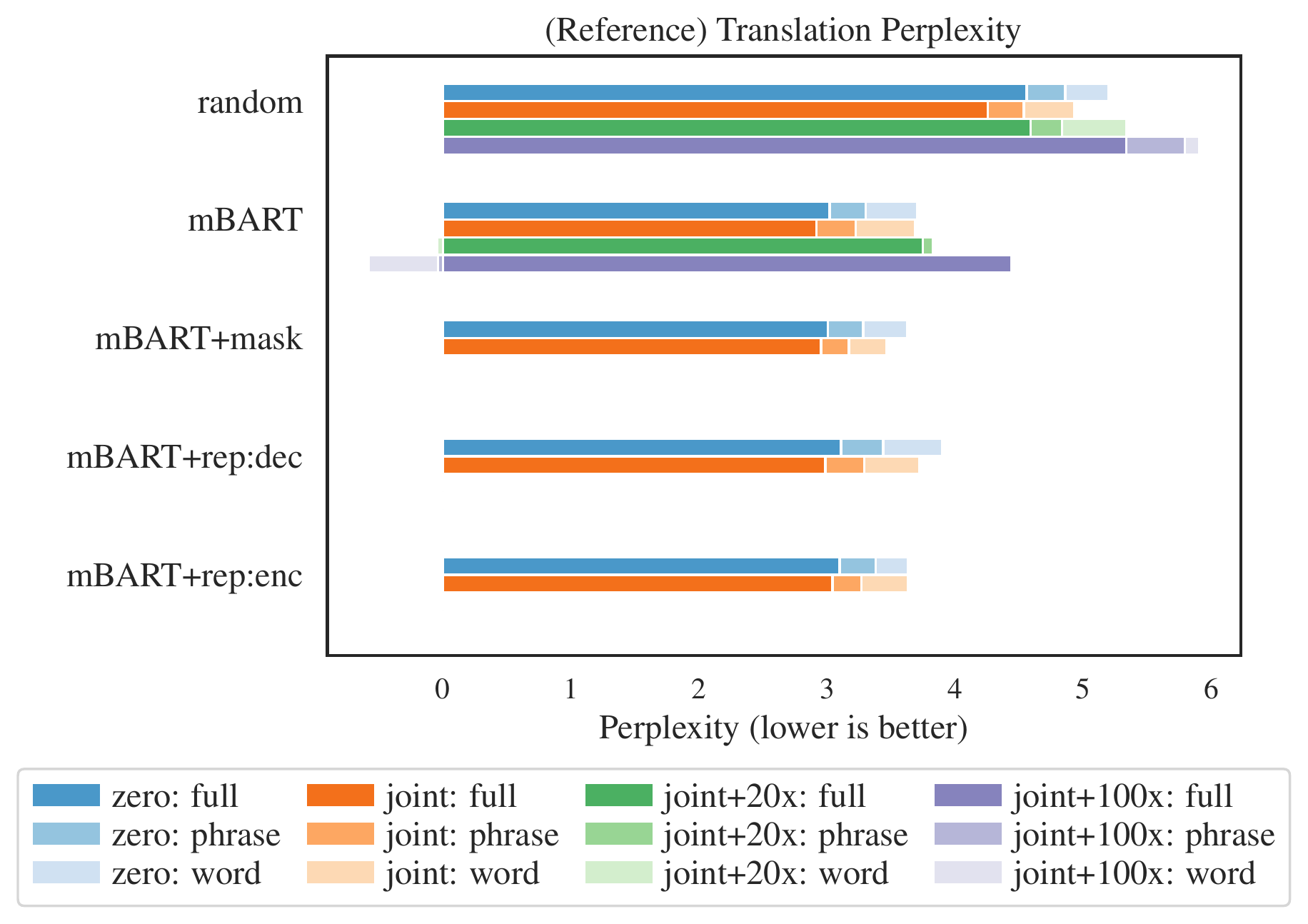}
  \caption{Results for the en$\rightarrow$fr models.}
  \label{fig:app-analysis-perplexity-enfr}
\end{subfigure}%
\hfill
\begin{subfigure}{.48\textwidth}
  \centering
  \includegraphics[width=1\columnwidth]{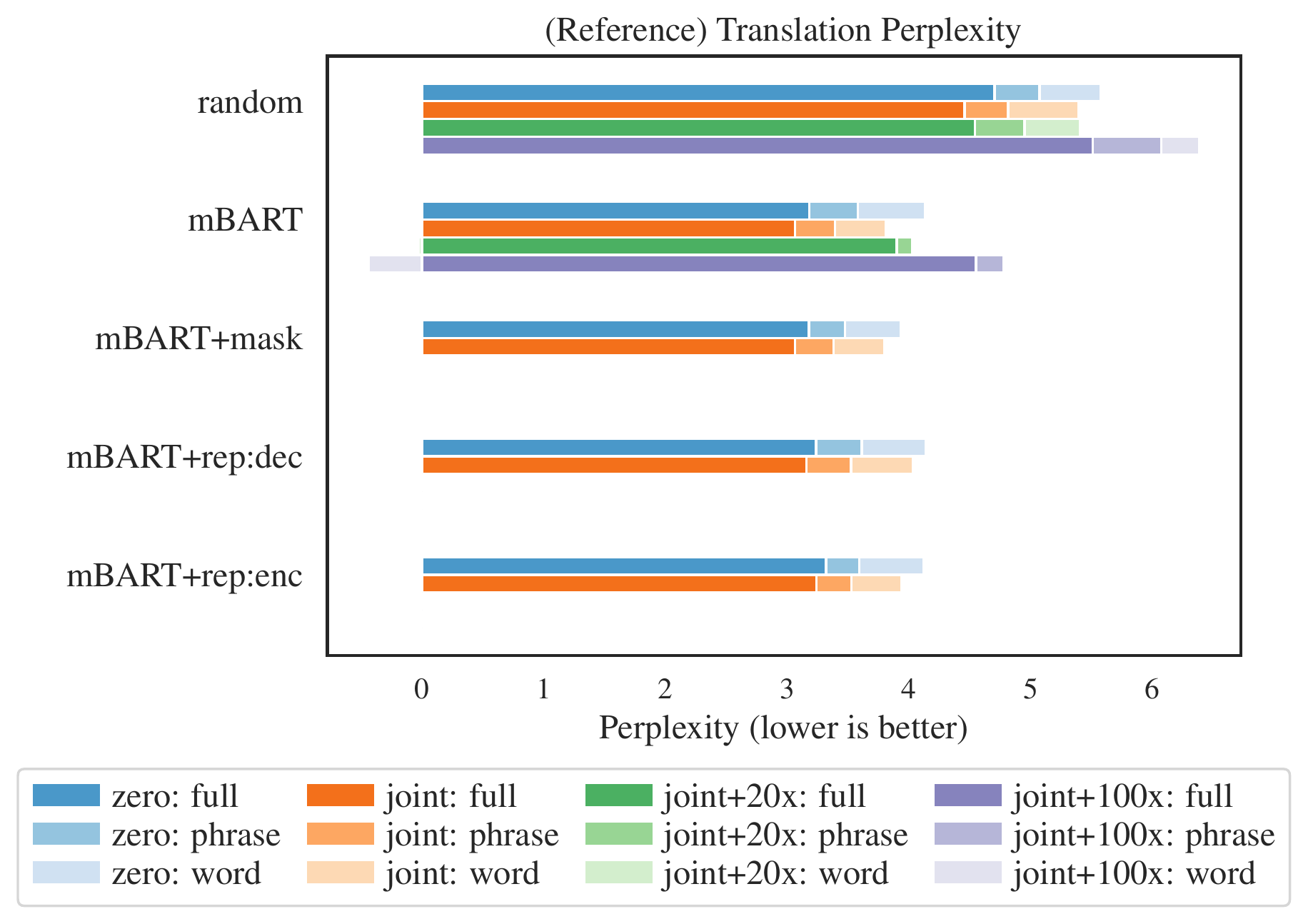}
  \caption{Results for the en$\rightarrow$es models.}
  \label{fig:app-analysis-perplexity-enes}
\end{subfigure}
    \caption{
    Variation in perplexity of reference translations, as we vary the idiom representations.
    The length of each (lighter) bar encodes the difference from its (darker) bar to its left.
    Negative bar lengths indicate a decrease relative to the (darker) bar before it.
    The darkest shades correspond to using the full (original) context for encoding idiom words, and each lighter shades correspond to narrower contexts.}
\label{fig:app-analysis-perplexity}
\end{figure*}

\begin{figure}[t]
    \centering
\includegraphics[width=1\columnwidth]{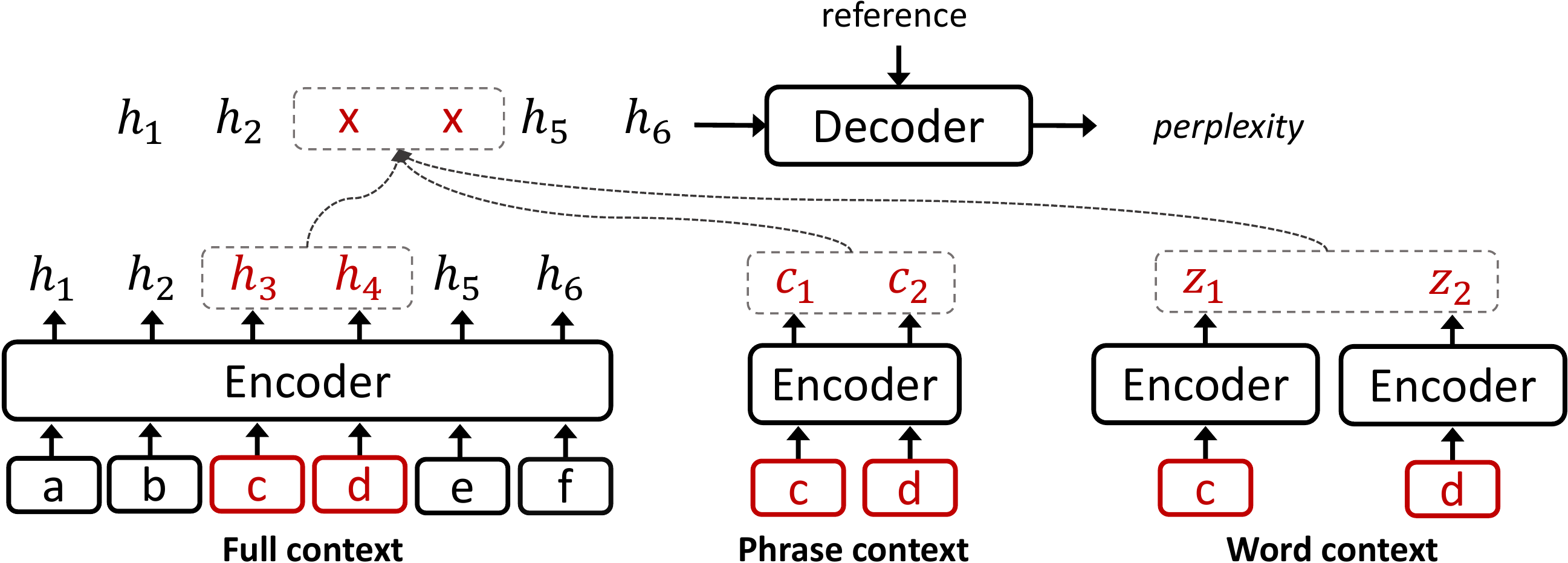}    
\caption{Illustration of how we measure the effects of different idiom contexts on translation likelihood (perplexity).
First, we obtain each sequence of idiom token representations, by encoding them within different contexts.
Then we feed each sentence to the encoder, and before passing its outputs to the decoder, 
we replace the idiom token representations with those whose context we want to probe.
We use the sequence of encoder outputs, to score the reference translation.
}
    \label{fig:app-analysis-likelihood}
\end{figure}

\subsection{Variation in Translation Likelihood}

With this probe, we test how the variation of the idiom encoder representations affects the likelihood of the reference translations.
Figure~\ref{fig:app-analysis-likelihood}, shows a visual example of how this probe works.
Specifically, we translate a sentence pair with teacher-forcing, 
but we replace the encoder idiom token representations, 
before passing them to the decoder, 
with the representations obtained after encoding them with different contexts. 
Finally, we measure the perplexity of the reference translation under the model. 
Figure~\ref{fig:app-analysis-perplexity}, compares models across splits and contexts.

Overall, the results are very consistent across both language pairs and all model variants.
We also observe that the behaviour of all the noisy finetuned mBART models very similar behaviour,
across all contexts.
However, there is a small increase in the perplexity assigned under the ``mBART+replace(dec)'' variant.
This is expected, as this model was trained with decoder dropout, which affect the LM capabilities of the model,
and consequently makes it assign an overall smaller probability to all sentences.
This increase in perplexity is observed in both language pairs, but is more pronounce in en$\rightarrow$fr.

The more we upsample the idiom-train sentence pairs, 
the less probable other sentences become under the model.
Surprisingly, 100x upsampling causes the pretrained model to yield lower perplexity with narrower context (i.e., lighter bars have negative length),
``reverting'' some of the effects of overfitting.
However, we don't have a satisfying explanation for this behaviour\footnotemark.
This phenomenon is observed in both language pairs.

\clearpage

\clearpage

\begin{figure*}[t]
\centering
\begin{subfigure}{.48\textwidth}
  \centering
  \includegraphics[width=1\columnwidth]{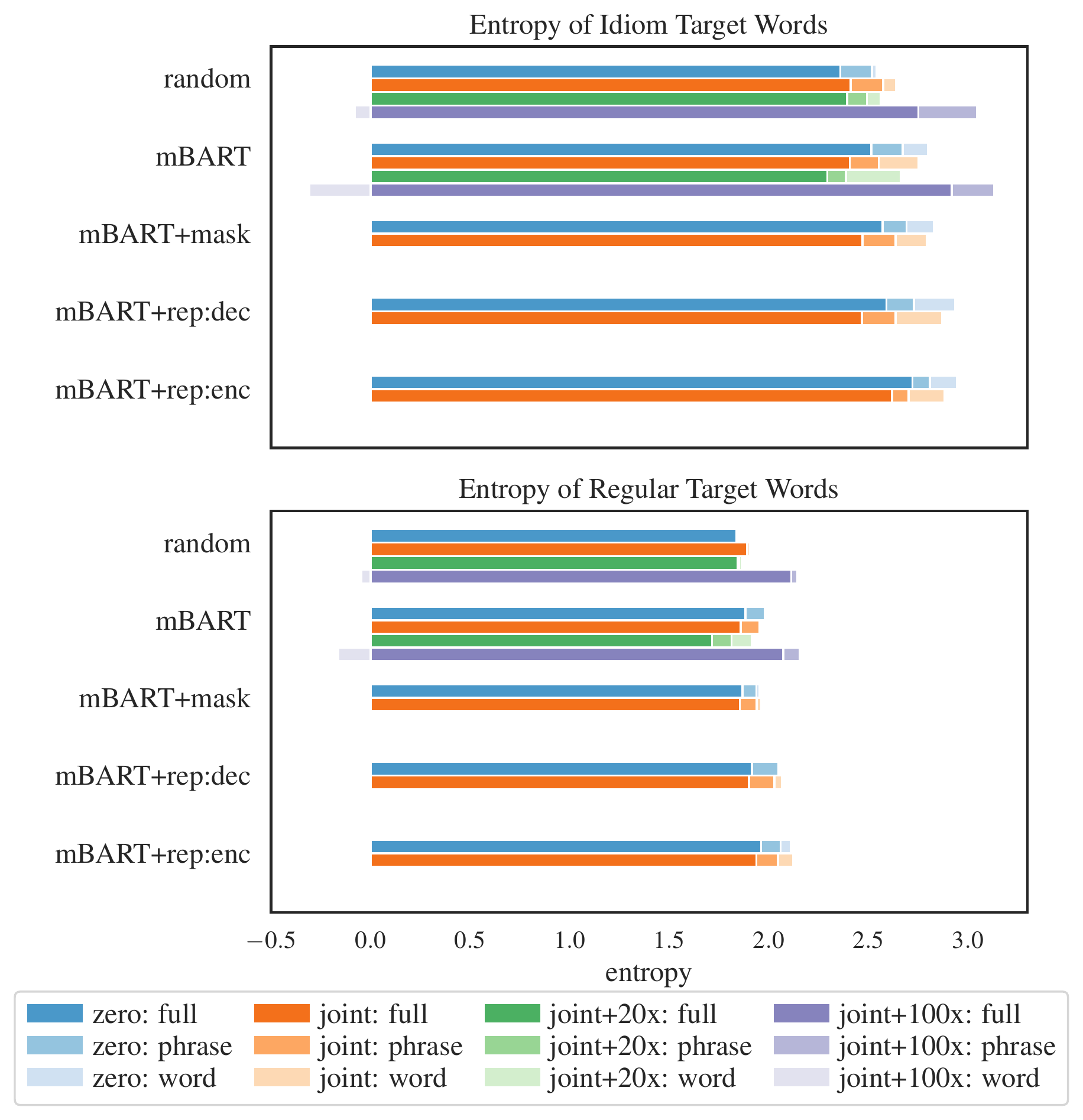}
  \caption{Results for the en$\rightarrow$fr models.}
  \label{fig:app-analysis-entropy-enfr}
\end{subfigure}%
\hfill
\begin{subfigure}{.48\textwidth}
  \centering
  \includegraphics[width=1\columnwidth]{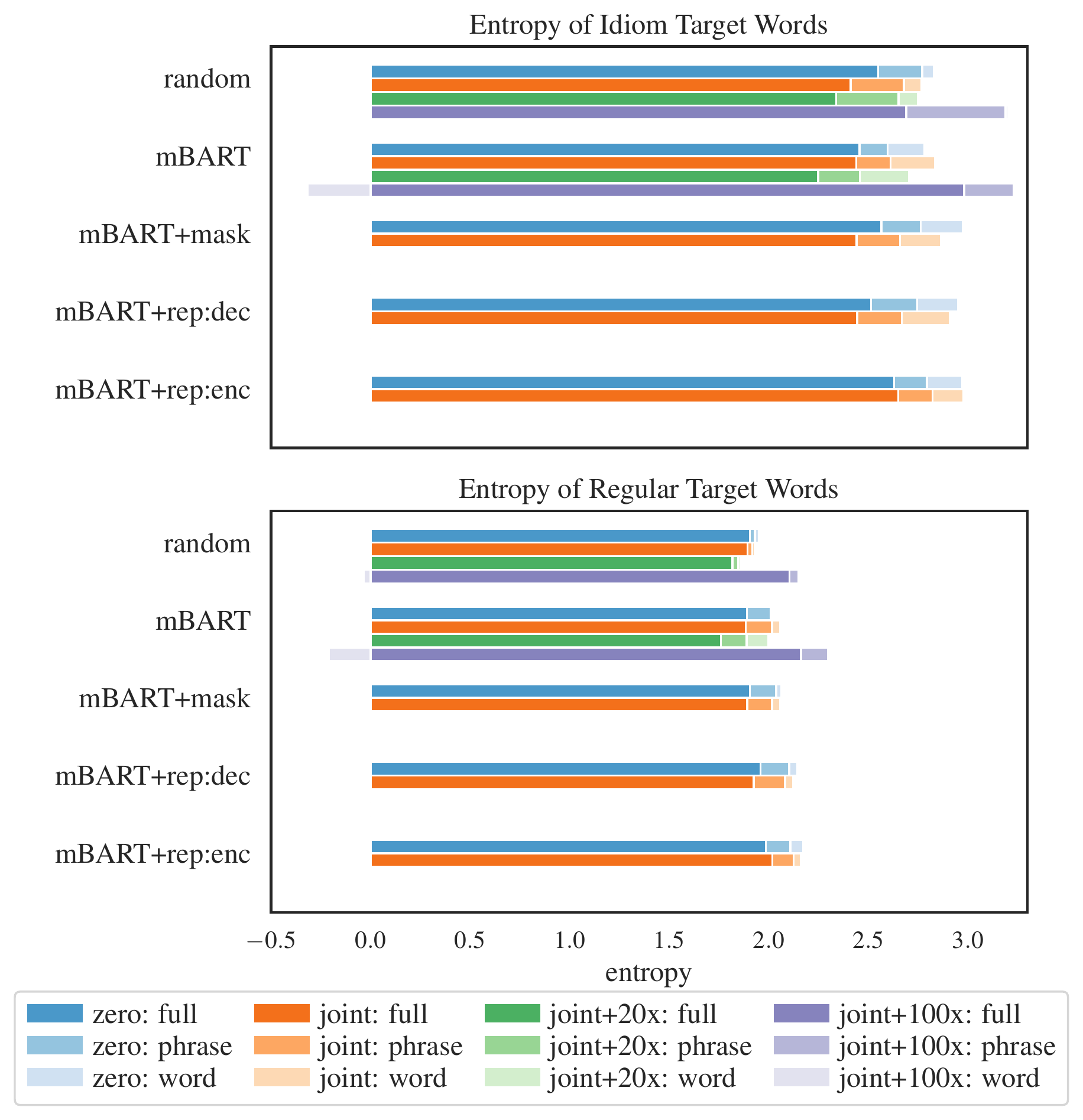}
  \caption{Results for the en$\rightarrow$es models.}
  \label{fig:app-analysis-entropy-enes}
\end{subfigure}
    \caption{
    Comparison of model uncertainty during the translation of regular-vs-idiom words.
    The length of each (lighter) bar encodes the difference from its (darker) bar to its left.
    Negative bar lengths indicate a decrease relative to the (darker) bar before it.
    }
\label{fig:app-analysis-entropy}
\end{figure*}

\begin{figure}[t]
    \centering
\includegraphics[width=1\columnwidth]{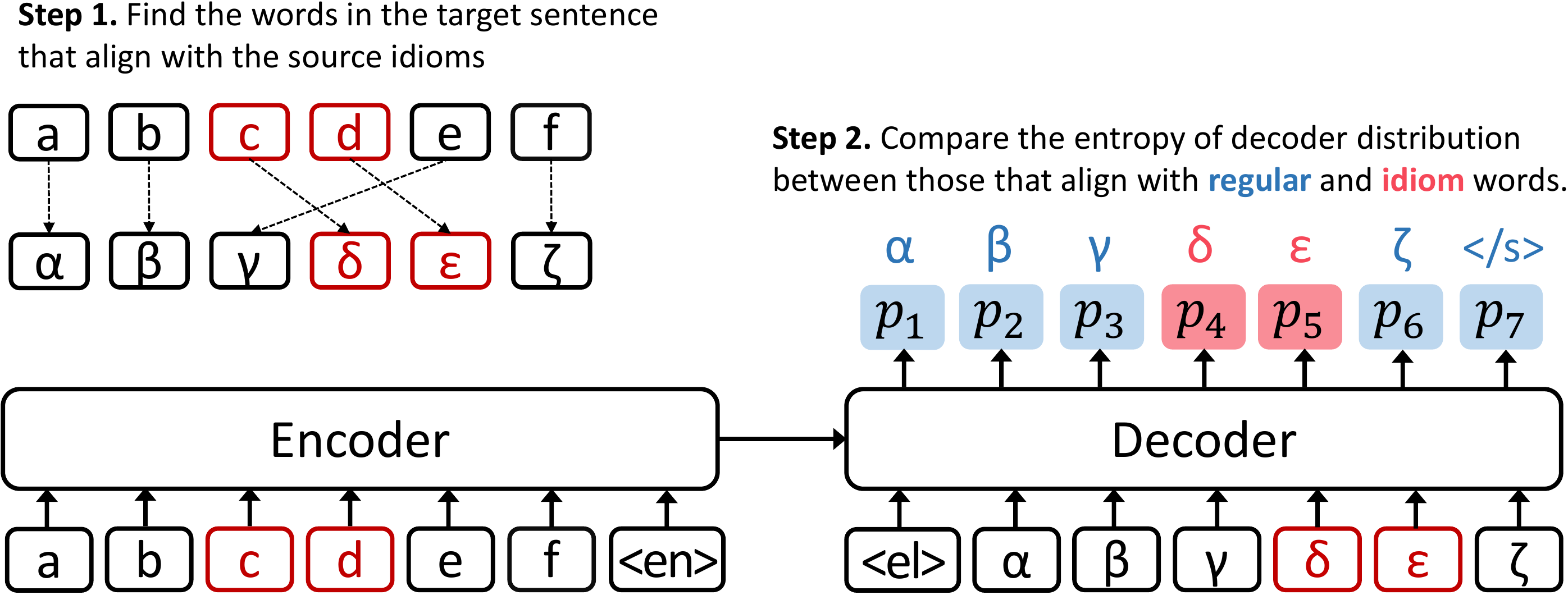}    
\caption{Illustration of how we compute the effects of different idiom contexts on the model'
s uncertainty.
First, we find which words in the target sentence align with the source idioms.
Then, we evaluate a sentence pair (i.e., teacher-forcing decoding), and compute the entropy of each token distribution produced by the decoder.
Using the word alignments, we compute the average entropy for words that translate idiom-vs-regular words.}
    \label{fig:app-analysis-uncertainty}
\end{figure}

\subsection{Decoder Uncertainty}
\label{sec:app-analysis-entropy}

Next, we focus on how the token-level uncertainty of the decoder varies while it translates 
idiom vs. non-idiom words (Figure~\ref{fig:analysis-entropy}).
For each model,
first, we translate each sentence pair with teacher-forcing\footnotemark 
and then measure the entropy of the decoder’s distributions for each target token.

Once more, this probe reveals that the models in both language pairs behave similarly.
As mentioned in the main paper, the distributions of words that translation the idiom phrase have significantly larger entropy that the rest.
This demonstrates that the models clearly are much more uncertain when translating idioms, even the pretrained ones.

Including and upsampling 20x the idiom-train data is helpful,
but extreme upsampling (i.e., 100x) is universally harmful,
although we observe a drop in uncertainty with word-level context. 
This is more pronounced when translating idiom words 
and suggests that models have overfitted to the words of the idiom phrases.

\clearpage

\end{document}